\documentclass{article}

\usepackage[preprint]{neurips_2026}

\usepackage[utf8]{inputenc}
\usepackage[T1]{fontenc}
\usepackage{fontawesome}
\usepackage{hyperref}
\hypersetup{
  pdftitle={GRAPE: Graph-Augmented Prototype Explanations for Interactive Medical Image Diagnosis},
  pdfauthor={Rasul Khanbayov, Erchin Serpedin, Hasan Kurban}
}
\usepackage{url}
\usepackage{booktabs}
\usepackage{amsfonts}
\usepackage{amsmath}
\usepackage{amssymb}
\usepackage{nicefrac}
\usepackage{microtype}
\usepackage{xcolor}
\usepackage{graphicx}
\usepackage{multirow}
\usepackage{algorithm}
\usepackage{algorithmic}
\usepackage{float}
\usepackage{subcaption}
\usepackage{tikz}
\usepackage{bm}
\usetikzlibrary{arrows.meta, positioning, shapes}
\title{GRAPE: Graph-Augmented Prototype Explanations for
Interactive Medical Image Diagnosis}

\author{%
  Rasul Khanbayov$^{1}$, Erchin Serpedin$^{2}$, Hasan Kurban$^{1,}$\thanks{Corresponding author: \texttt{hkurban@hbku.edu.qa}}\\
  $^{1}$ College of Science and Engineering, Hamad Bin Khalifa University, Doha, Qatar\\
  $^{2}$ Department of Computer and Electrical Engineering,\\
  Texas A\&M University, College Station, TX, USA
}
\begin{document}

\maketitle

\begin{abstract}
Prototype-based medical image classifiers present three clinical limitations: they treat findings as independent, silently amplify unsafe physician feedback, and require full retraining whenever a new finding is needed. We present GRAPE (Graph-Augmented Prototype Explanations), a unified architecture that targets all three challenges.
First, a Graph Attention Task Head boosts macro-F1 by +13.8\,pp over the prototype baseline on TBX11K but not on NIH ChestX-ray14, with no graph-specific benefit established. Second,
a Concept-Mismatch Safety Check, to our knowledge the first such mechanism in prototype-based medical classifiers, warns when the model's dominant finding inside a doctor-drawn region conflicts with the claimed label, catching 85\% of simulated erroneous annotations with no extra inference cost at a favorable operating point, though sensitivity falls substantially at clinically realistic false-positive budgets. Third, Open-Vocabulary Prototype Anchoring aligns visual prototypes to clinical text, allowing a new finding to be added from a single labeled image without modifying any other component, evaluated for localization only. On NIH ChestX-ray14, one Effusion example matched that concept's low fully-supervised Pointing Game ceiling in one run but not in a second. On TBX11K, prototype maps achieve $2.1{\times}$ better lesion localization than end-to-end concept-based baselines. The graph head adds only $+1$~ms latency at interactive batch size. The project page is available at \url{https://kurbanintelligencelab.github.io/GRAPE}.
\end{abstract}

%% ─────────────────────────────────────────────────────────────────
\section{Introduction}
\label{sec:intro}
%% ─────────────────────────────────────────────────────────────────

Deep learning has achieved radiologist-level accuracy on several medical imaging tasks~\cite{rajpurkar2017chexnet,liu2023revisiting}, yet clinical adoption remains limited. Two barriers stand out. First, black-box models provide no evidence for their predictions, making it impossible for a physician to verify or override the reasoning. Second, even ``interpretable'' models may silently propagate spatial errors when doctors provide corrective feedback, creating a patient-safety risk.

Prototype-based models~\cite{chen2019looks,donnelly2022deformable} address the first barrier by making predictions through similarity to learned visual examples. Recent work on Concept-based Similarity Reasoning (CSR)~\cite{huy2025interactive} extends this to multi-prototype per-concept reasoning with doctor-in-the-loop interaction. However, three structural limitations remain:

\begin{enumerate}
    \item \textbf{Independent concepts.} The task head is a flat linear layer applied to concatenated concept similarity scores. It ignores known anatomical correlations between findings (e.g., active tuberculosis (TB) and cavity formation co-occur reliably), discarding clinically meaningful structure.
    \item \textbf{Unsafe spatial interaction.} When a physician  draws a bounding box to guide the model's attention, the system applies the spatial feedback regardless of the model's confidence in that region. If the model is confused and associates a different concept with the indicated region, the feedback amplifies an error.
    \item \textbf{Closed concept vocabulary.} Adding a new clinical finding (e.g., a rare variant not seen in training) requires fully retraining the prototype atlas, a prohibitive cost in clinical settings.
\end{enumerate}

We propose \textbf{GRAPE}, a unified architecture that targets all three limitations within a coherent four-stage training framework. Our contributions are:

\begin{itemize}
    \item A \textbf{Graph Attention Task Head} (Module A, §\ref{sec:gnn}) that constructs a data-driven concept co-occurrence graph from training labels and refines prototype similarity scores via a two-layer GAT-style head before classification. In the evaluated implementation, the graph does not affect the head's outputs at inference.
    \item A \textbf{Concept-Mismatch Safety Check} (Module B, §\ref{sec:uncertainty}), to our knowledge the first such mechanism in prototype-based medical classifiers that computes per-concept mean similarity inside a physician-drawn region and warns when a different concept dominates, flagging potential error amplification before feedback is applied.
    \item \textbf{Open-Vocabulary Prototype Anchoring} (Module C, §\ref{sec:vlm}) that aligns visual prototypes to frozen BioViL-T text embeddings, enabling a new clinical concept to be added at test time from a single labeled image without modifying any other component.
    \item A \textbf{four-stage training pipeline} (§\ref{sec:training}) that integrates concept supervision, contrastive prototype learning, VLM alignment, and graph-structured classification.
\end{itemize}

%% ─────────────────────────────────────────────────────────────────
\section{Related Work}
\label{sec:related}
%% ─────────────────────────────────────────────────────────────────

ProtoPNet~\cite{chen2019looks} learns class-specific visual prototypes. Extensions add deformability~\cite{donnelly2022deformable}, decision trees~\cite{nauta2021neural}, part-sharing~\cite{rymarczyk2020protopshare}, and patch variants~\cite{nauta2023pip}. MProtoNet~\cite{wei2023mprotonet} applies prototypes to brain tumor MRI, and Ma et al.~\cite{ma2023protoconcepts} visualize prototypical concepts with multiple images. GRAPE builds on CSR~\cite{huy2025interactive}, replacing its linear head with a graph-structured classifier. Concept Bottleneck Models~\cite{koh2020concept,espinosa2022concept,yuksekgonul2023posthoc} interpose concept scores before prediction. Margeloiu et al.~\cite{margeloiu2021concept} show they can be semantically misaligned, so GRAPE grounds prototypes to BioViL-T~\cite{bannur2023learning} text anchors instead. GCN classifiers~\cite{chen2019multi,ye2020attention} model corpus-derived label co-occurrence. GRAPE uses training-label statistics with prototype-similarity node features and learned GAT attention~\cite{velivckovic2017graph}. Bayesian uncertainty~\cite{gal2016dropout,lakshminarayanan2017simple} requires sampling overhead. GRAPE's prototype-variance signal is deterministic~\cite{sensoy2018evidential}. Medical VLMs (MedCLIP~\cite{wang2022medclip}, BiomedCLIP~\cite{zhang2023biomedclip}) target classification. GRAPE uses text embeddings as geometric anchors for prototype initialization, preserving spatial interpretability. Interactive annotation methods~\cite{wang2018interactive} lack concept-mismatch checks. Attribution quality is compared via Insertion/Deletion AUC~\cite{petsiuk2018rise}, following the causal deletion/insertion measures it introduces, without implying our own method's maps are shown to be more faithful. B-cos networks~\cite{bohle2022bcos} offer an alternative faithful attribution approach by replacing standard convolutions with alignment-based transforms. Unlike GRAPE, they do not support interactive concept-level correction or open-vocabulary extension.

%% ─────────────────────────────────────────────────────────────────
\section{Method}
\label{sec:method}

\subsection{Preliminaries: Prototype-Based Concept Reasoning}
\label{sec:prelim}

Backbone $F$ followed by two-layer MLP projector $P$ yields L2-normalized spatial embeddings:
\begin{equation}
    \mathbf{f}' = P(\mathbf{f}) \in \mathbb{R}^{D \times h \times w}, \quad \|\mathbf{f}'_{:,i,j}\|_2 = 1 \;\;\forall i,j.
\end{equation}

Each concept $k$ has $M$ L2-normalized prototypes $\{\mathbf{p}_{km}\} \subset \mathbb{R}^D$. The cosine similarity map $\mathcal{S}_{km}(i,j) = \langle \mathbf{p}_{km}, \mathbf{f}'_{:,i,j} \rangle$ aggregates to a scalar score:
\begin{equation}
    s_{km} = \max_{i,j}\, \mathcal{S}_{km}(i,j).
    \label{eq:sim_score}
\end{equation}

The baseline CSR model feeds the score matrix $\mathbf{s} \in \mathbb{R}^{K \times M}$ through a linear task head to produce class logits. GRAPE replaces this with the three modules described below.

% \vspace{-2mm}
\subsection{Module A: Graph Attention Task Head}
\label{sec:gnn}

Clinical concepts are not independent. Active TB and cavity formation co-occur, and cardiomegaly correlates with pleural effusion, yet a flat linear head over concatenated $s_{km}$ discards this structure. We build a data-driven co-occurrence graph from training labels $\mathbf{y}^{(n)} \in \{0,1\}^K$. The $K \times K$ conditional probability matrix

\begin{equation}
    \mathbf{A}_{kk'} = P(k \mid k') = \frac{\sum_{n} y^{(n)}_k \cdot y^{(n)}_{k'}}{\sum_{n} y^{(n)}_{k'} + \epsilon}
\end{equation}

is thresholded at $\tau$ ($0.1$ on TBX11K, $0.05$ on NIH). Edges with $\mathbf{A}_{kk'}>\tau$ keep their weight, self-loops get weight one, and the weights are row-normalized and scale the GAT attention logits. The matrix is \emph{not} symmetrized, so $\mathbf{A}_{kk'} \neq \mathbf{A}_{k'k}$ in general. We denote the resulting graph by $G$.
% Since $\mathbf{y}^{(n)}\in\{0,1\}^K$, the numerator counts co-occurrences of $k$ and $k'$, the denominator counts occurrences of $k'$, and the shared $1/N$ factor cancels to yield the empirical $P(k\mid k')$~\cite{chen2019multi}; $\epsilon=1$ (denominator clamped to $\min=1$) prevents division by zero for concepts absent from training.
Since $\mathbf{y}^{(n)}$ is binary, the numerator counts samples where both concepts occur and the denominator counts samples where $k'$ occurs, so $\mathbf{A}_{kk'}$ is the maximum-likelihood estimate of $P(k\mid k')$~\cite{chen2019multi}.
The constant $\epsilon$ (denominator clamped to $\min=1$) prevents division by zero for concepts absent from training.

\noindent
\textbf{Graph Attention Network.} Each concept is a node with initial feature $\mathbf{h}_k^{(0)} = \mathbf{s}_{k,:} \in \mathbb{R}^M$. Two GAT layers~\cite{velivckovic2017graph} refine these via learned attention (Eqs.~\ref{eq:gat_e}--\ref{eq:gat_h}). A LayerNorm–Dropout–Linear readout produces class logits. Layer dimensions and head counts are in Table~\ref{tab:hparams_fixed}. \textbf{Implementation note.} The evaluated head departs from Eq.~\ref{eq:gat_h}: each node aggregates its own transformed features, so with fixed weights its outputs do not depend on $G$ as long as every node keeps an incoming edge. $G$ affects training only through attention dropout.

\begin{figure}[t]
  \centering
  \includegraphics[width=0.99\linewidth]{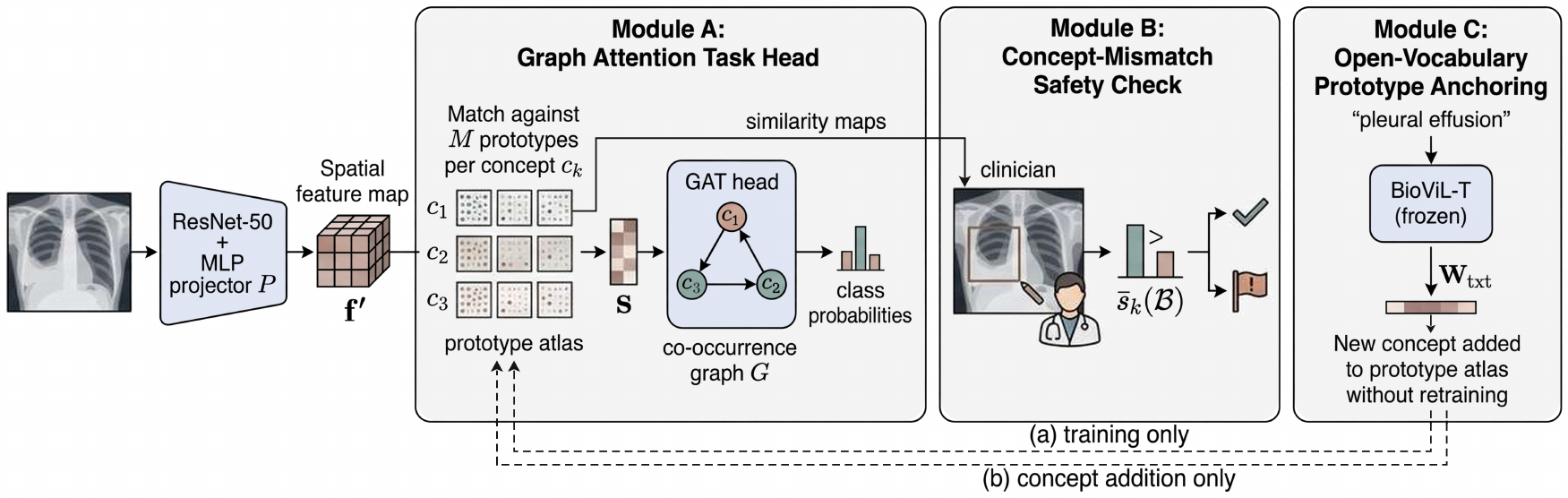}
\caption{\textbf{GRAPE architecture.} An X-ray is encoded by ResNet-50 + MLP projector $P$ into spatial features $\bm{f}'$. \textbf{Module A} matches $\bm{f}'$ against a learned prototype atlas, and score matrix $\mathbf{s}$ feeds a GAT-style head built over a co-occurrence graph $G$ to produce per-class probabilities. \textbf{Module B} compares mean concept similarity $\bar{s}_k(\mathcal{B})$ inside a clinician's box against the claimed concept's, flagging a mismatch before feedback is applied. \textbf{Module C} adds concepts from a text anchor via a frozen BioViL-T encoder projected through $\bm{W}_{\!\text{txt}}$. Dashed arrows mark training-only (a) or concept-addition-only (b) paths.}
  \label{fig:multiclass}
\end{figure}

\subsection{Module B: Concept-Mismatch Safety Check}
\label{sec:uncertainty}

\paragraph{Prototype variance maps.}
The disagreement among $M$ prototypes for concept $k$ at patch $(i,j)$ provides a spatial consistency estimate without sampling:
\begin{align}
    \mu_k(i,j) &= \frac{1}{M} \sum_{m=1}^M \mathcal{S}_{km}(i,j), \\
    U_k(i,j) &= \frac{1}{M} \sum_{m=1}^M \!\left(\mathcal{S}_{km}(i,j) - \mu_k(i,j)\right)^2.
    \label{eq:uncertainty}
\end{align}
High $U_k(i,j)$ indicates that the $M$ prototypes strongly disagree about whether patch $(i,j)$ belongs to concept $k$, signaling ambiguous evidence. The \emph{safety check} uses the mean map $\mu_k$ directly to detect concept mismatch. $U_k$ serves only as an optional gate.
\paragraph{Spatial safety check.}
When a doctor provides a bounding box $\mathcal{B}$ intending to reinforce concept $k^*$, GRAPE checks whether the dominant concept inside the region matches.
It first computes the mean similarity per concept inside $\mathcal{B}$:
\begin{equation}
    \bar{s}_k(\mathcal{B}) = \frac{1}{|\mathcal{B}|} \sum_{(i,j) \in \mathcal{B}} \mu_k(i,j) \quad \forall k,
\end{equation}
identifies the dominant concept $k^\dagger = \arg\max_k \bar{s}_k(\mathcal{B})$, and issues a safety warning if $k^\dagger \neq k^*$ and $\bar{s}_{k^\dagger}(\mathcal{B}) - \bar{s}_{k^*}(\mathcal{B}) > \eta$ (threshold $\eta = 0.05$). The warning is raised before the feedback is applied.
This \emph{concept-mismatch} check is designed to catch cases where the model's dominant response inside the region conflicts with the claimed concept label, before the error can be amplified in the prototype atlas. The rule is uncertainty-free because it compares $\bar{s}_k(\mathcal{B})$ across concepts and never reads $U_k$.
We use the spatial mean rather than the spatial max of $\mu_k$ inside $\mathcal{B}$, as the two perform within noise of each other (Table~\ref{tab:mean_vs_max}).
\paragraph{Optional $U_k$-gate.} $U_k$ (Eq.~\ref{eq:uncertainty}) is used at inference time as an \emph{optional second-stage filter} applied only to boxes the dominance rule already flagged. A flagged mismatch is suppressed if the claimed concept's own per-prototype variance $U_{k^*}(\mathcal{B})$ (the variance across its $M$ prototypes of their mean similarity inside $\mathcal{B}$) exceeds a gate threshold $\theta_V$ (distinct from $\eta$ above). This gate trades true-positive detection for false-positive suppression (§\ref{sec:safety_eval}).
% Figure~\ref{fig:interaction} illustrates this interaction workflow.

% \begin{figure}[t]
%   \centering
%   \includegraphics[width=\linewidth]{figures/figure_C_interaction.png}
%   \caption{\textbf{GRAPE spatial interaction demo.} \emph{Left:} Active TB similarity map before interaction — the max activation (orange square) falls outside the GT bounding box (blue). \emph{Centre:} A doctor draws a positive bounding box (\texttt{bb+}) around the lesion region. \emph{Right:} After applying the importance-weighted interaction map (Eq.~12), the model's attention is redistributed toward the annotated region, improving localisation. The uncertainty map (Module B) is checked before this step to ensure the box region is not dominated by a conflicting concept.}
%   \label{fig:interaction}
% \end{figure}

\subsection{Module C: Open-Vocabulary Prototype Anchoring}
\label{sec:vlm}

We eliminate the closed-vocabulary bottleneck by anchoring each prototype to a clinical text description via the frozen BioViL-T encoder~\cite{bannur2023learning}. A learnable projection $\mathbf{W}_{\text{txt}} \in \mathbb{R}^{D \times 768}$ maps each concept's text embedding $\mathbf{t}_k$ into the visual prototype space:
\begin{equation}
    \hat{\mathbf{t}}_k = \frac{\mathbf{W}_{\text{txt}}\mathbf{t}_k}{\|\mathbf{W}_{\text{txt}}\mathbf{t}_k\|_2} \in \mathbb{R}^D.
\end{equation}
The \emph{alignment loss} (Stage 3) pulls all $M$ prototypes toward their text anchor while keeping the BioViL-T backbone frozen:
\begin{equation}
    \mathcal{L}_\text{align} = \frac{1}{KM} \sum_{k=1}^K \sum_{m=1}^M \left(1 - \langle \mathbf{p}_{km},\, \hat{\mathbf{t}}_k \rangle\right).
    \label{eq:align_loss}
\end{equation}
\textbf{Zero-shot concept addition.} At test time, a clinician adds a new concept $k'$ without image annotations by initializing $M$ perturbed prototypes from the text anchor ($\sigma_{\text{init}}=0.1$):
\begin{equation}
    \mathbf{p}_{k'm}^{(0)} = \frac{\hat{\mathbf{t}}_{k'} + \sigma_{\text{init}}\boldsymbol{\epsilon}_m}{\|\hat{\mathbf{t}}_{k'} + \sigma_{\text{init}}\boldsymbol{\epsilon}_m\|_2}, \quad \boldsymbol{\epsilon}_m \sim \mathcal{N}(\mathbf{0}, \mathbf{I}_D),\; \boldsymbol{\epsilon}_m \leftarrow \boldsymbol{\epsilon}_m / \|\boldsymbol{\epsilon}_m\|_2.
\end{equation}
The perturbation keeps the $M$ prototypes distinct. Without it they would be identical and receive identical gradients during fine-tuning. These prototypes are immediately usable for inference or adaptable with a small number of labeled examples via concept-local fine-tuning.

\subsection{Training Pipeline}
\label{sec:training}

\begin{figure}[!htbp]
  \centering
  \includegraphics[width=0.88\linewidth]{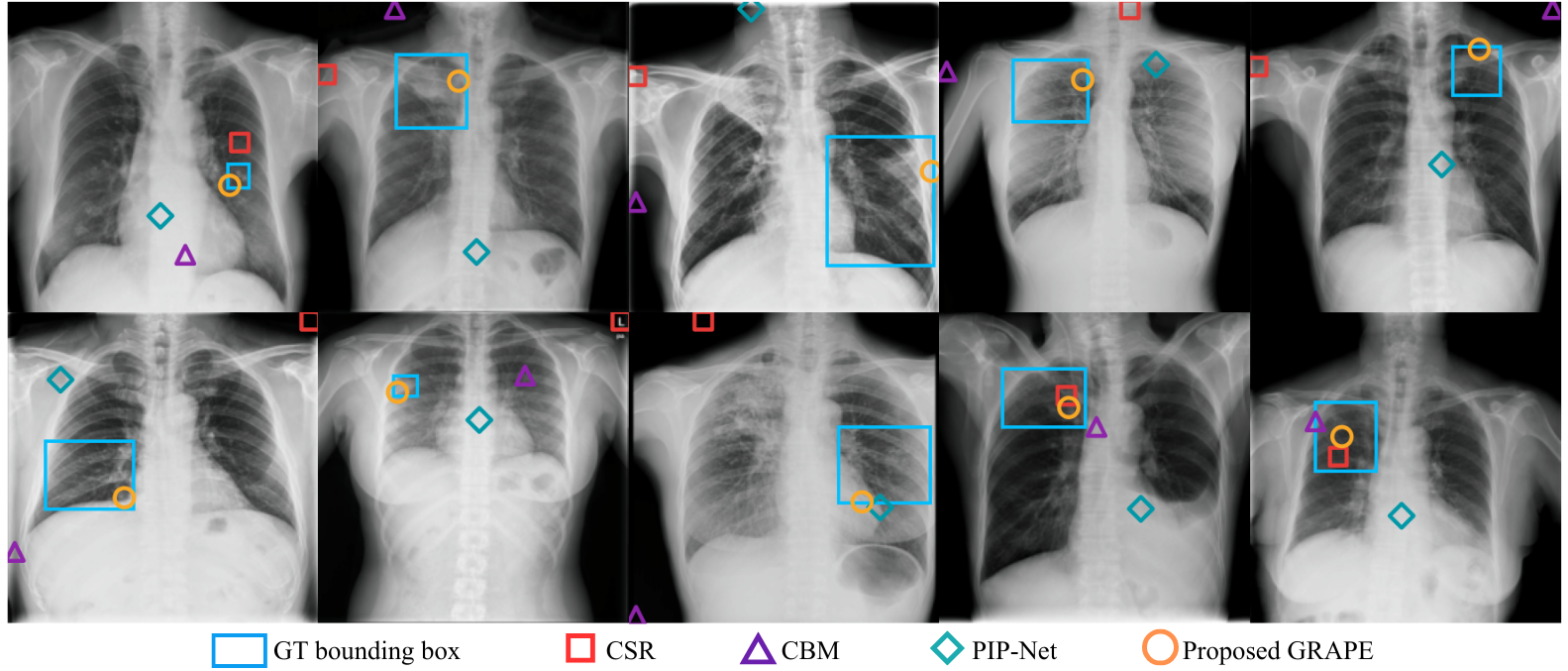}
  \caption{\textbf{Pointing Game on TBX11K chest X-rays (Active TB concept).} Each marker shows the max-activation point of a method. In these examples, \textbf{GRAPE} (orange circle) localizes the lesion inside the ground-truth bounding box (cyan), while the other methods' maxima mostly fall outside it. These images were selected because \textbf{GRAPE}'s point is a hit.}
  \label{fig:pg_dots}
\end{figure}

GRAPE trains in four stages (Algorithm~\ref{alg:grape}): concept supervision, concept vector generation, prototype learning with optional VLM alignment, and GNN task head training.

\textbf{Stage 1: Concept Supervision.} The backbone $F$ and per-concept CAM heads are trained with BCE on Global Average Pooled activations (Eq.~\ref{eq:l1}).

\noindent
\textbf{Stage 2: Concept Vector Generation.} CAM-weighted spatial averages of frozen backbone features produce per-concept vectors $\mathbf{v}_k$ (Eq.~\ref{eq:camvec}) that seed the Stage 3 prototype atlas.

\noindent
\textbf{Stage 3: Prototype Learning with VLM Alignment.}
The projector $P$, prototypes $\{\mathbf{p}_{km}\}$, and text projection $\mathbf{W}_{\text{txt}}$ are trained jointly. Each projected concept vector $\mathbf{v}'_k = P(\mathbf{v}_k)$ serves as a query, and the multi-prototype contrastive loss assigns it to its ground-truth concept $\tilde{k}$ via soft assignment:
\begin{align}
    q_m^k(\mathbf{v}') &= \text{softmax}_m\!\left(\gamma \langle \mathbf{p}_{km},\mathbf{v}' \rangle\right), \\
    \text{sim}_k(\mathbf{v}') &= \sum_m q_m^k(\mathbf{v}') \langle \mathbf{p}_{km},\mathbf{v}' \rangle, \\
    \mathcal{L}_\text{con} &= -\log \frac{\exp\!\left(\lambda(\text{sim}_{\tilde{k}}(\mathbf{v}'_{\tilde{k}}) + \delta)\right)}{\sum_{k} \exp\!\left(\lambda\, \text{sim}_k(\mathbf{v}'_{\tilde{k}})\right)},
    \label{eq:contrastive}
\end{align}
with $\lambda=10$, $\gamma=5$, $\delta=0.1$. As written and implemented, $\delta$ enters only the numerator, so it offsets the loss by a constant and does not change its gradients. The Stage~3 loss is $\mathcal{L}_3 = \mathcal{L}_\text{con} + 0.1\,\mathcal{L}_\text{align}$. Our implementation applies this weight twice, so the effective weight is $0.01$.

\noindent 
\textbf{Stage 4: Graph-Structured Classification.} With all prior weights frozen (batch-norm statistics still update), the GNN is trained with $\mathcal{L}_4 = \text{CE}(\text{GNN}(\mathbf{s}, G),\, y)$.

\begin{figure}[!htbp]
    \centering
    \includegraphics[width=0.90\linewidth]{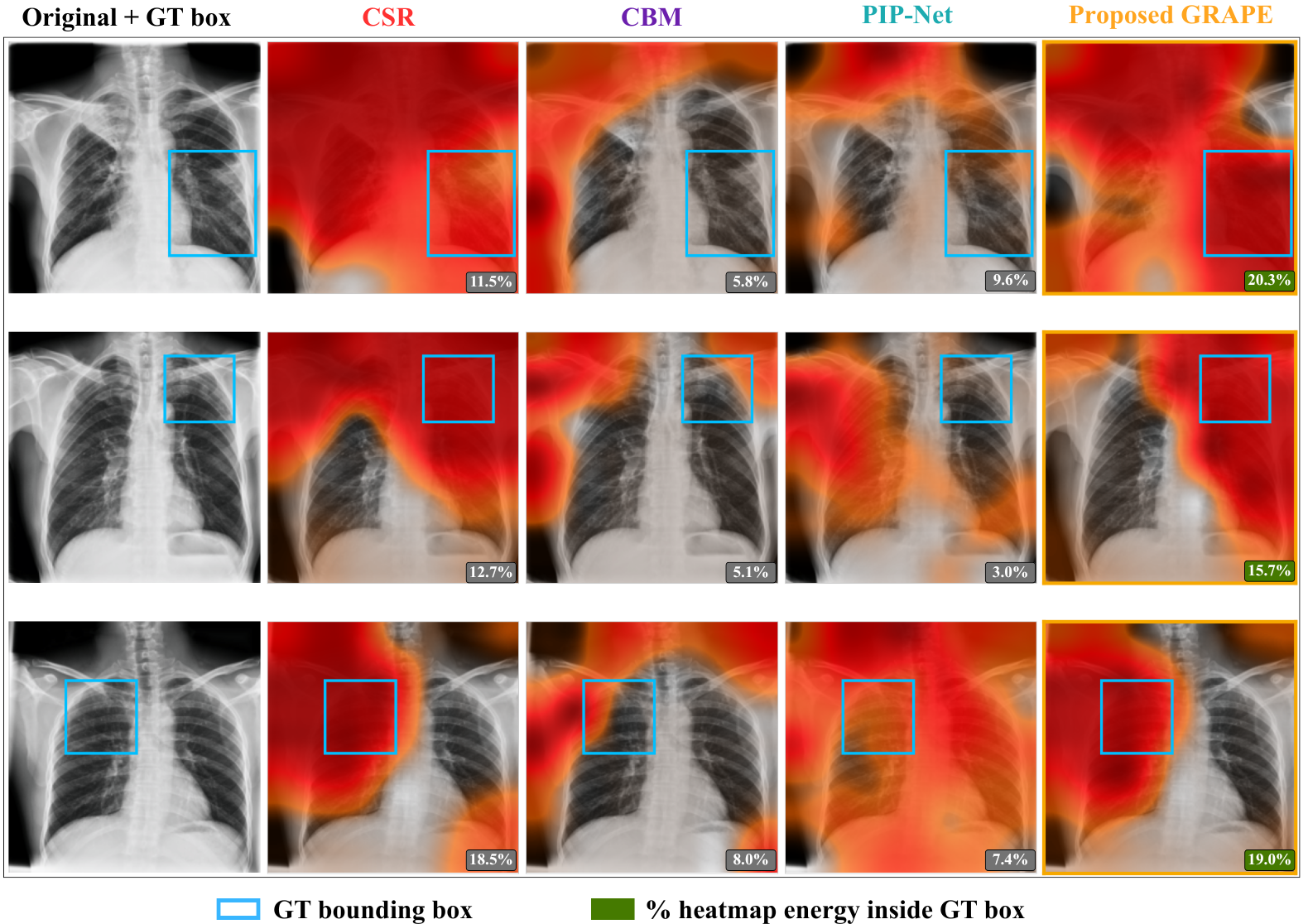}
    \caption{\textbf{Prototype similarity maps for CSR Baseline, CBM, PIP-Net, and GRAPE on three Active TB examples.} Heatmaps are overlaid on the input X-rays. Cyan boxes mark ground-truth lesions, and percentage badges give the fraction of total heatmap energy inside the box. Images were chosen to maximize \textbf{GRAPE}'s box-energy advantage and are not representative.}
    \label{fig:simmaps_cmp_panel}
\end{figure}

% \begin{figure}[!htbp]
%   \centering

%   \begin{subfigure}[t]{0.88\linewidth}
%     \centering
%     \includegraphics[width=\linewidth]{figures/pg_dots_18_modified_3.png}
%     \caption{Pointing Game localization.}
%     \label{fig:pg_dots_panel}
%   \end{subfigure}

%   \vspace{0.75em}

%   \begin{subfigure}[t]{0.98\linewidth}
%     \centering
%     \includegraphics[width=\linewidth]{figures/simmaps_comparison-overlay.png}
%     \caption{Prototype similarity map comparison.}
%     \label{fig:simmaps_cmp_panel}
%   \end{subfigure}

%   \caption{\textbf{Qualitative localization analysis on TBX11K Active TB chest X-rays.}
%   \textbf{(a)} Pointing Game visualization: each marker denotes the maximum-activation location of a method relative to the ground-truth lesion box (cyan). Aggregate hit rates across all 200 \texttt{bbox\_eval} images are reported in Table~\ref{tab:tbx11k} and Appendix~\ref{app:pg_perconcept}.
%   \textbf{(b)} Prototype similarity maps for CSR Baseline, CBM, PIP-Net, and GRAPE across three representative Active TB examples. Heatmaps are overlaid on the input X-rays; cyan boxes denote ground-truth lesion annotations; percentage badges report the fraction of total heatmap energy inside the ground-truth box. Aggregate box-energy statistics are consistent with the Pointing Game results in Table~\ref{tab:tbx11k}.}
%   \label{fig:tbx11k_qualitative_localization}
% \end{figure}
\vspace{-1mm}
%% ─────────────────────────────────────────────────────────────────
\section{Experiments}
\label{sec:experiments}
Tables numbered 4 and above and Figure~\ref{fig:proto_dispersion} are in the appendix.
%% ─────────────────────────────────────────────────────────────────
\vspace{-1mm}
\subsection{Datasets}
\label{sec:datasets}

\textbf{TBX11K}~\cite{liu2023revisiting} is a 3-class TB chest X-ray dataset. We use its 8,400 labeled images (3,800 healthy, 3,800 sick-but-non-TB, and 800 active-TB), with 1,211 bounding box annotations for two TB finding types (Active Tuberculosis, Obsolete Pulmonary Tuberculosis); the third, Pulmonary Tuberculosis, has no boxes or positive labels. We use the official 6,600/1,800 train/val split. Since TBX11K does not release test labels, the 1,800-image validation set also serves as the test split for F1 reporting, and the Stage-4 checkpoint is selected on it. The \texttt{bbox\_eval} split (200 active-TB images with at least one annotated box) is a subset of the same set and is used for Pointing Game evaluation. All methods share this protocol.

\textbf{NIH ChestX-ray14}~\cite{wang2017chestx} contains 100k+ frontal chest X-rays with 14 finding labels. Primary multi-seed comparison (Table~\ref{tab:compare}) uses a stratified patient-level sub-sample of 10k patients (5 seeds, no patient leakage). Single-seed ablations (Table~\ref{tab:nih}) use a 20k-image shard (seed~42). Pointing Game uses 709 shard images with 888 image--concept pairs (964 boxes) from BBox\_List\_2017.\footnote{BBox\_List\_2017 has 880 annotated images. Boxes were placed on each patient's first streamed image, so NIH Pointing Game and Module~B values are approximate.}

\noindent
\textbf{Evaluation metrics.} \textbf{Macro F1} measures classification performance averaged across classes, accounting for class imbalance. \textbf{Pointing Game (PG)}~\cite{zhang2018top} evaluates spatial interpretability. For each (image, concept) pair with a ground-truth bounding box, it checks whether the spatial maximum of the concept's similarity map falls inside the box, and we report the hit rate across all evaluated pairs.
Hyperparameters are listed in Tables~\ref{tab:hparams_fixed} and~\ref{tab:hparams_dataset}.

% \begin{figure}[t]
%   \centering
%   \includegraphics[width=\linewidth]{figures/figure_B_similarity_maps.png}
%   \caption{\textbf{GRAPE per-concept similarity maps and uncertainty for a single Active TB image.} Left: input chest X-ray with GT bounding box (blue). Columns 2--4: similarity maps for each of the three TB-type concepts — the Active TB map fires strongly on the upper-lobe lesion region, while the other concepts show low activation. Right: prototype variance (uncertainty) map for Active TB — low variance in the lesion region indicates the 100 prototypes agree, validating the prediction. Orange square marks the max activation point; green = inside GT box (HIT).}
%   \label{fig:simmaps}
% \end{figure}

\subsection{Main Results}
\label{sec:results}

\begin{table}[!htbp]
  \caption{Ablation on \textsc{TBX11K} (3-class TB detection). We report Macro F1
  and Pointing Game (PG); higher is better for both. All results are
  mean\,$\pm$\,std over three seeds $\{0,1,42\}$. GRAPE is designed to
  improve grounding (PG) under a curriculum protocol and is \emph{not}
  expected to surpass end-to-end baselines on F1 (Observation~1). Best F1 and
  PG within each block are in \textbf{bold}. The CSR macro-F1 ($0.706$) differs
  from Tables~\ref{tab:compare} ($0.728$), \ref{tab:graph_heads} ($0.7519$), and
  \ref{tab:n6_mlp} ($0.798$) because each uses a different seed count or
  protocol (see Table~\ref{tab:n6_mlp}).}
  \label{tab:tbx11k}
  \centering
  \small
  \setlength{\tabcolsep}{6pt}
  \renewcommand{\arraystretch}{1.15}
  \begin{tabular}{@{}lccccc@{}}
    \toprule
    \multirow{2}{*}{Method} & \multicolumn{3}{c}{Module} &
    \multirow{2}{*}{Macro F1\,$\uparrow$} & \multirow{2}{*}{PG\,$\uparrow$} \\
    \cmidrule(lr){2-4}
    & GNN & Unc.\textsuperscript{\S} & VLM & & \\
    \midrule
    \multicolumn{6}{@{}l}{\textit{Module ablation under curriculum training}} \\
    CSR\textsuperscript{\textdagger}~\cite{huy2025interactive}
      & --         & --         & --
      & $0.706\,{\pm}\,0.059$ & $0.077\,{\pm}\,0.016$ \\
    \quad + GNN
      & \checkmark & --         & --
      & $0.845\,{\pm}\,0.049$ & $0.117\,{\pm}\,0.045$ \\
    \quad + VLM
      & --         & --         & \checkmark
      & $0.722\,{\pm}\,0.032$ & $0.097\,{\pm}\,0.044$ \\
    \quad + GNN\,+\,VLM
      & \checkmark & --         & \checkmark
      & $\mathbf{0.875\,{\pm}\,0.020}$ & $0.087\,{\pm}\,0.033$ \\
    \textbf{GRAPE} (GNN\,+\,Unc.)
      & \checkmark & \checkmark & --
      & $0.845\,{\pm}\,0.049$ & $\mathbf{0.117\,{\pm}\,0.045}$ \\
    \midrule
    \multicolumn{6}{@{}l}{\textit{End-to-end variants and concept-based baselines}\textsuperscript{\textparagraph}} \\
    ResNet-50 (backbone only)\textsuperscript{\textbullet}
      & --         & --         & --
      & $\mathbf{0.993}$ & -- \\
    GRAPE-E2E\textsuperscript{\textdaggerdbl}
      & \checkmark & \checkmark & --
      & $0.9900\,{\pm}\,0.0020$ & $\mathbf{0.0910\,{\pm}\,0.0590}$ \\
    CBM~\cite{koh2020concept}
      & --         & --         & --
      & $0.9778\,{\pm}\,0.0109$ & $0.0550\,{\pm}\,0.0187$ \\
    PIPNet~\cite{nauta2023pip}
      & --         & --         & --
      & $0.9644\,{\pm}\,0.0021$ & $0.0533\,{\pm}\,0.0184$ \\
    \bottomrule
  \end{tabular}
  \begin{flushleft}\footnotesize
  \textsuperscript{\textdagger}\,CSR re-implemented; no published numbers on these splits.\quad
  \textsuperscript{\S}\,Module~B is inference-only; does not affect F1/PG.\quad
  \textsuperscript{\textdaggerdbl}\,E2E: task gradients through backbone throughout.\quad
  \textsuperscript{\textbullet}\,ResNet-50+GAP+linear, seed~42; E2E gains driven by unconstrained fine-tuning.\quad
  \textsuperscript{\textparagraph}\,Same backbone/splits; CBM: joint bottleneck; PIPNet: patch-prototype similarity.
  \end{flushleft}
\end{table}

\paragraph{Key observations.}
\textbf{(1) The nonlinear task head is the dominant module on TBX11K.} The GNN head alone saturates the staged gain (+13.9\,pp F1), while VLM alone yields no improvement on the sparse $K=3$ concept set. E2E training reaches $0.990 \pm 0.002$ F1 but at the cost of localization (PG $0.091 \pm 0.059$ vs.\ staged $0.111 \pm 0.038$, Table~\ref{tab:n4}). A plain ResNet-50 linear head achieves $0.993$ F1, confirming E2E gains are driven by unconstrained backbone fine-tuning rather than concept reasoning.
The per-class gain is concentrated on majority classes (Healthy $+21.5$\,pp, Sick-non-TB $+20.6$\,pp, Table~\ref{tab:perclass_f1}). Active~TB F1 is comparable between GRAPE and CSR, suggesting the GNN reshapes the output distribution rather than improving disease-specific detection.

\textbf{(2) VLM alignment on NIH.} VLM contributes to the single-seed gain (Table~\ref{tab:nih}), but multi-seed evaluation (Table~\ref{tab:compare}) shows GRAPE and CSR are statistically comparable ($p=0.894$), so the gain does not persist under the patient-level split protocol.

\textbf{(3) Pointing Game localization.} GRAPE's prototype maps achieve $\mathbf{2.1\times}$ \textbf{higher PG than CBM} ($0.117$ vs.\ $0.055$) \textbf{and PIPNet} ($0.117$ vs.\ $0.053$, 3-seed means, Table~\ref{tab:tbx11k}). End-to-end fine-tuning sacrifices prototype-to-region alignment for classification accuracy. On the single-seed run (seed~42), GRAPE achieves PG~$=0.1401$ vs.\ CSR $0.0628$ ($+7.73$~pp, Table~\ref{tab:graph_heads}). Multi-seed evaluation finds no statistically significant difference vs.\ CSR ($p=0.64$), reflecting high variance ($\sigma \approx 0.03$) inherent to Pointing Game at $7{\times}7$ resolution.

\subsection{Graph Head Comparison and Inference Speed}
\label{sec:graph_heads}

We ablate the choice of graph task head by comparing GRAPE's GAT with ML-GCN~\cite{chen2019multi} and ADD-GCN~\cite{ye2020attention}. All three heads outperform the linear baseline (+10.7--14.6\,pp F1) and perform within seed variance of one another (Table~\ref{tab:graph_heads}), so the choice of head is not the driver of the gain. We retain the GAT-style head as the evaluated configuration.

\textbf{Graph structure vs.\ added nonlinearity.} All three graph heads are nonlinear readouts, so that comparison alone cannot separate a relational (graph) effect from a capacity (nonlinearity) effect. To isolate the two, we train a parameter-matched 2-layer MLP (${\approx}47$K parameters, Table~\ref{tab:n6_mlp}) directly on the flattened $K{\times}M$ score vector, replacing graph structure with a generic nonlinear function of the same capacity. On TBX11K, the MLP achieves F1~$=0.865$, closing $\sim\!69\%$ of the gap between the linear head ($0.798$) and the GAT ($0.896$) on this single seed. The remaining $+3.1$\,pp ($0.865 \to 0.896$) is a difference between readouts, not a graph effect: the sole cross-concept edge is worth $0.25$\,pp (Table~\ref{tab:edge_ablation}), and the evaluated head's outputs do not depend on $G$ at inference. The same comparison on NIH ($K=14$, 5 seeds, patient-level split, Table~\ref{tab:n6_nih}) gives MLP~$=0.5849\pm0.0189$ vs.\ GAT~$=0.6050\pm0.0065$ (+2.01\,pp, paired $t(4)=2.33$, $p=0.080$). Here the MLP has 188K parameters vs.\ the GAT's 46K, since its input grows with $K$. The GAT does not clear the MLP by more than seed variance at conventional significance, so no graph-specific benefit is established on either dataset. The GAT did not lose on any of the 5 seeds and showed substantially lower variance.

We also compare a parameter-matched (${\approx}48$K) self-attention head with no predefined adjacency (Table~\ref{tab:n13_selfattn}). On TBX11K it reaches F1~$=0.906$, PG~$=0.085$, comparable to the GAT's F1~$=0.900$, PG~$=0.091$ on the same seed (a separate run from Table~\ref{tab:n6_mlp}). A nonlinear readout therefore performs comparably with or without a predefined adjacency, consistent with a nonlinear-readout effect rather than a graph effect at this concept count.
A per-edge ablation (Table~\ref{tab:edge_ablation}) shows self-loops dominate on TBX11K ($K=3$), since removing a node's only incoming edge silences it. On NIH ($K=14$, 105 edges), no graph-specific benefit is established.
The GNN head adds only $+1$~ms over the linear head at batch size~1, satisfying real-time interaction requirements (Table~\ref{tab:speed}).

% \vspace{-2mm}
\begin{table}[!htbp]
  \caption{Ablation on NIH ChestX-ray14 (binary No Finding vs.\ any finding,
  14 concepts), seed~42 of the initial development run on a 20k-image
  streaming shard. Module configurations match those in
  Table~\ref{tab:tbx11k}. Multi-seed mean\,$\pm$\,std on the
  patient-level split is reported in Table~\ref{tab:compare}; per-metric
  winner in \textbf{bold}.}
  \label{tab:nih}
  \centering
  \small
  \setlength{\tabcolsep}{8pt}
  \renewcommand{\arraystretch}{1.2}
  \begin{tabular}{@{}lcccccc@{}}
    \toprule
    \multirow{2}{*}{Method} & \multicolumn{3}{c}{Module}
      & \multirow{2}{*}{Macro F1\,$\uparrow$}
      & \multirow{2}{*}{PG\,$\uparrow$} \\
    \cmidrule(lr){2-4}
    & GNN & Unc. & VLM & & \\
    \midrule
    CSR~\cite{huy2025interactive}
      & --         & --         & --         & $0.6166$           & $\mathbf{0.0766}$ \\
    \quad + GNN\,+\,Unc.
      & \checkmark & \checkmark & --         & $0.6526$           & $0.0664$           \\
    \quad + Unc.\,+\,VLM
      & --         & \checkmark & \checkmark & $0.5859$           & $0.0743$           \\
    GRAPE (full)
      & \checkmark & \checkmark & \checkmark & $\mathbf{0.7138}$  & $0.0642$           \\
    \bottomrule
  \end{tabular}
  % \begin{flushleft}\footnotesize
  % \textsuperscript{\textdagger}\,Re-implemented under our training protocol
  %   and hyperparameters; published CSR numbers on these splits are
  %   unavailable for direct comparison.
  % \end{flushleft}
\end{table}

\subsection{Comparison with CSR Baseline}

Table~\ref{tab:compare} summarizes GRAPE vs.\ CSR~\cite{huy2025interactive} over five seeds on both datasets (patient-level split for NIH). On TBX11K, GRAPE's F1 ranges 0.798--0.916 across seeds (std = 0.042), and the paired $t$-test yields $t(4)=8.03$, $p=0.0013$.

\begin{table}[!htbp]
  \caption{Summary comparison of GRAPE against the CSR baseline under the
  \emph{same} interpretability-constrained training regime (frozen backbone,
  concept supervision, staged curriculum). All results are mean\,$\pm$\,std
  over five seeds $\{0,1,2,3,42\}$, with significance from a paired $t$-test on
  per-seed Macro F1. End-to-end fine-tuned methods (CBM, PIPNet) use a
  different training paradigm and are compared in Table~\ref{tab:sota}.
  Tables~\ref{tab:tbx11k} and~\ref{tab:nih} give full ablations. This table's CSR
  macro-F1 on TBX11K ($0.728$, 5-seed average) differs from other protocols'
  measurements of the same quantity (see Table~\ref{tab:n6_mlp}).}
  \label{tab:compare}
  \centering
  \small
  \setlength{\tabcolsep}{6pt}
  \renewcommand{\arraystretch}{1.2}
  \begin{tabular}{@{}llcccc@{}}
    \toprule
    Dataset & Method & Macro F1\,$\uparrow$ & PG\,$\uparrow$
            & $\Delta$F1 & $p$-value \\
    \midrule
    \multirow{2}{*}{TBX11K}
      & CSR\textsuperscript{\textdagger}
        & $0.728\,{\pm}\,0.050$ & $0.094\,{\pm}\,0.033$ & --     & --     \\
      & GRAPE (GNN\,+\,Unc.)
        & $\mathbf{0.866\,{\pm}\,0.042}$ & $0.080\,{\pm}\,0.032$\textsuperscript{\P}
        & $\mathbf{+13.8\,\text{pp}}$ & $\mathbf{0.0013}$ \\
    \midrule
    \multirow{2}{*}{NIH CXR14}
      & CSR\textsuperscript{\textdagger}
        & $0.6048\,{\pm}\,0.0170$ & --\textsuperscript{\S} & --   & --     \\
      & GRAPE (full)
        & $0.6031\,{\pm}\,0.0174$ & --\textsuperscript{\S}
        & $-0.2$\,pp & $0.894$ (n.s.) \\
    \bottomrule
  \end{tabular}
  \begin{flushleft}\footnotesize
  \textsuperscript{\textdagger}\,CSR is re-implemented under our training
    protocol and hyperparameters; published numbers on these splits are
    unavailable for direct comparison.\quad
  \textsuperscript{\S}\,NIH bounding-box annotations were unavailable for
    the multi-seed evaluation run; single-seed PG is reported in
    Table~\ref{tab:nih}.\quad
  \textsuperscript{\P}\,Paired $t$-test on per-seed PG: $t(4)=-0.50$, $p=0.64$ (n.s.). The
    development-run single-seed advantage ($+7.73$\,pp, Table~\ref{tab:graph_heads}) does not
    replicate consistently across seeds (seed-level range: $-10.5$ to $+6.0$\,pp).
  \end{flushleft}
\end{table}
% \vspace{-2mm}
%% ─────────────────────────────────────────────────────────────────
\subsection{Module B: Safety Check Evaluation}
\label{sec:safety_eval}
%% ─────────────────────────────────────────────────────────────────

We simulate miscorrections on the TBX11K \texttt{bbox\_eval} split (207 image--concept pairs, one box per pair, from 309 boxes). With probability $p$, a bounding box is reassigned to a wrong concept, and the safety check (the dominance rule following Eq.~6, $\eta=0.05$) issues a warning when the dominant concept inside the box differs from the claimed one. This is simulated disagreement detection, a controllable proxy for model-clinician disagreement, not demonstrated interactive safety under real user behavior. Bootstrap 95\% CIs use $B=1000$ resamples.

With the $U_k$-gated variant ($\theta_V = 2{\times}10^{-4}$) as the default operating point, Module B catches \textbf{85.1\%} of misdrawn boxes with an FP rate of \textbf{16.4\%} [12.1\%, 20.9\%] (Table~\ref{tab:safety_gated}). This operating point is selected by minimizing FP subject to TP~$\geq 0.85$ on this same split. $\theta_V$ is checkpoint-specific and must be re-selected per checkpoint. The ungated baseline (Table~\ref{tab:safety}) achieves higher TP (88--89\%) at the cost of a higher FP rate (21.7\% [16.4\%, 27.1\%]). Tightening to $\theta_V=10^{-4}$ drops FP to 12.6\% but TP falls to 78.0\%, below the 85\% clinical threshold. At clinically realistic false-positive budgets the true-positive rate falls substantially further, to $55.7\%$ TP at $7.7\%$ FP and $44.5\%$ TP at $3.4\%$ FP (Table~\ref{tab:safety_gated}). We use the $U_k$-gated configuration as the default here and report the low-false-positive regime as an explicit limitation.
Table~\ref{tab:n5} shows that the ungated dominance rule substantially outperforms warnings triggered by MC-Dropout uncertainty (TP 90.5\% vs.\ 51.0\%/23.4\% at FP 22--25\%) while requiring only a single deterministic forward pass.
On NIH ($K=14$), TP remains high ($0.921\pm0.012$) but FP rises to $0.731\pm0.011$, possibly due to dense concept overlap and approximate box placement. At low false-positive budgets on NIH, a single global threshold reaches $17.9\%$ TP at $4.6\%$ FP and $20.1\%$ TP at $5.4\%$ FP (Table~\ref{tab:nih_lowfp}). Module B's sensitivity at clinically realistic false-positive rates is therefore substantially lower on NIH than on TBX11K, consistent with the denser concept overlap at $K=14$ and suggesting that larger vocabularies need per-concept thresholds, not a single global one.

%% ─────────────────────────────────────────────────────────────────
\subsection{Further Analysis: Zero-Shot Concept Addition and Attribution}
\label{sec:zeroshot_eval}
%% ─────────────────────────────────────────────────────────────────

\textbf{Module C: Zero-shot concept addition.} We evaluate Module C via a leave-one-concept-out experiment on TBX11K ($K=3$). For each held-out concept, we reinitialize its $M=100$ prototypes with Eq.~9 and fine-tune them with $n \in \{0,1,5,20\}$ labeled images, measuring Pointing Game on the \texttt{bbox\_eval} split (Table~\ref{tab:zeroshot}). This run loads the model without its text aligner, so Eq.~9 falls back to a random unit vector. Table~\ref{tab:random_anchor} compares the BioViL-T text anchor with a random one.
With no labeled images, this initialization yields non-zero localization for \textit{active\_tb} (PG~$=0.030$), and a single labeled image reaches PG~$=0.068 \pm 0.045$, on average 41\% of full supervision (68\% CI: 14--69\%). With five seeds the few-shot trend is essentially monotonic, with 1-shot (PG~$=0.068$) $<$ 5-shot (PG~$=0.079$) $\approx$ 20-shot (PG~$=0.078$).

We also evaluate on NIH ChestX-ray14 ($K=14$) with the same fallback, where 3 of 4 target concepts have bbox coverage (Infiltration, Effusion, Atelectasis). Effusion's 1-shot PG matched full supervision in this run (PG~$=0.072$, a low absolute ceiling) but not in a second run, and Infiltration's 20-shot reaches 53\% of full supervision (Table~\ref{tab:zeroshot_nih}). Adding a concept is not free. Classification F1 on NIH drops by up to 6.7\,pp and dominant-concept assignments flip in 19--46\% of other findings' boxes (Appendix~\ref{app:zeroshot_nih}), so prototype-level non-interference does not imply decision-level non-interference.

A matched random-anchor control (Table~\ref{tab:random_anchor}) tests whether the BioViL-T text anchor itself, rather than any prototype initialization, is responsible for this localization. The two TBX11K concepts disagree about which anchor wins, and with only three seeds neither result is conclusive. The control therefore does not establish one-example localization above chance (Appendix~\ref{app:random_anchor}), or that the text-anchoring mechanism specifically makes it possible.

\textbf{Similarity map attribution.}\label{sec:faithfulness} We compare Insertion and Deletion AUC~\cite{petsiuk2018rise}, which measure how quickly predicted class probability changes as pixels ranked by saliency are removed or revealed, for GRAPE and CSR on the 200 active-TB images of the TBX11K test split (Table~\ref{tab:faithfulness}). GRAPE's Deletion AUC ($0.954\pm0.068$) and Insertion AUC ($0.886\pm0.148$) versus CSR's ($0.923\pm0.104$ and $0.873\pm0.138$) differ by less than one standard deviation on both metrics. We report this as comparable attribution behavior, not as evidence that either method's similarity maps are more faithful to the underlying decision. Table~\ref{tab:sota} gives a contextual comparison with SymFormer, CheXNet, CBM, and PIPNet.
% (Table~\ref{tab:sota}).
\vspace{1mm}
%% ─────────────────────────────────────────────────────────────────
%% ─────────────────────────────────────────────────────────────────
\section{Discussion}
\label{sec:discussion}
%% ─────────────────────────────────────────────────────────────────

VLM alignment over-regularizes sparse concept sets ($K=3$). After Stage~3, all $M=100$ prototypes cluster tightly near their text anchors (Figure~\ref{fig:proto_dispersion}, mean prototype–anchor cosine $=0.995\pm0.008$ for TBX11K vs.\ $0.992\pm0.007$ for NIH), and inter-prototype pairwise cosines are similarly high ($0.990$ vs.\ $0.984$), confirming that the alignment loss collapses intra-concept prototype diversity in both settings.

\textbf{Safety implications of the concept-mismatch check.} The dominance rule following Eq.~6 does not improve macro-F1 directly. Its value is in flagging potentially harmful interactions. In a deployment scenario where a doctor mistakenly draws a box over a region dominated by a different finding, the safety warning gives the system a chance to flag the inconsistency before the feedback is applied.
\paragraph{Limitations.}
Pointing Game scores remain low in absolute terms (7–14\%) because similarity maps at $7 \times 7$ resolution must be compared to bounding boxes annotated in 224×224 space. A finer-resolution backbone is untested, and an in-domain backbone at the same $7{\times}7$ resolution is compared in Table~\ref{tab:backbone_ablation}. Modules A, B, and C all operate on the prototype similarity score matrix rather than backbone features directly, so a stronger in-domain backbone is expected to change the quality of the scores these modules receive rather than what the modules do with them. Module B's true-positive rate falls substantially at clinically realistic false-positive budgets (§\ref{sec:safety_eval}). The safety-check evaluation itself is a controllable proxy for model-clinician disagreement, simulated via label corruption. It is not a validated model of real clinician behavior under ambiguous boxes, partial lesions, multi-finding regions, or varying clinician expertise. The deployment-relevant disagreement rate that would determine whether any given false-positive rate is tolerable requires a user study we do not have. Module~C is additionally evaluated on NIH ($K=14$, 3 concepts with bbox coverage). NIH boxes were placed on each patient's first streamed image, so NIH Pointing Game and Module~B values are approximate. Full $K=14$ leave-one-out and a clinician user study remain future work. Robustness of the static co-occurrence graph to a shifted disease prevalence has not been established, because the evaluated head's outputs do not depend on the graph at inference (Table~\ref{tab:distshift}).

\textbf{Training variance.} TBX11K F1 ranges $0.798$--$0.916$ (std $=0.042$, 5 seeds), reflecting Stage~4 GNN sensitivity on the $n{=}6{,}600$ dataset.

%% ─────────────────────────────────────────────────────────────────
\section*{Broader Impact}
\label{sec:impact}
%% ─────────────────────────────────────────────────────────────────

By making every prediction traceable to a specific visual concept and prototype, GRAPE enables clinicians to verify or override model reasoning rather than accepting opaque outputs.
The concept-mismatch safety check is designed to reduce the risk that erroneous clinician feedback silently degrades the prototype atlas, a failure mode that could otherwise go undetected in real deployments.
The open-vocabulary extension lowers barriers to adoption in resource-limited settings by allowing new clinical concepts to be introduced without full retraining. At the same time, prototype-based explanations that appear clinically plausible may not be faithful to the model's actual decision boundary, and clinicians who over-rely on similarity maps risk anchoring diagnostic judgment to potentially misleading visualizations.
This work does not evaluate performance across demographic subgroups. Prototype atlases learned from imbalanced distributions may exhibit worse spatial alignment for underrepresented populations, and practitioners should audit subgroup performance before deployment.

\paragraph{Ethics statement.}
This work uses two publicly released datasets: TBX11K~\cite{liu2023revisiting}, released under a research license, and NIH ChestX-ray14~\cite{wang2017chestx}, released by the U.S.\ National Institutes of Health (NIH) under a public-access license.
No new human-subject data were collected for this research.
All experiments are conducted retrospectively on de-identified images.
The system described in this paper is a research prototype and is \emph{not} deployed in any clinical setting.

%% ─────────────────────────────────────────────────────────────────
\section{Conclusion}
\label{sec:conclusion}
%% ─────────────────────────────────────────────────────────────────

We presented GRAPE, an interactive medical image diagnosis architecture combining concept-mismatch interaction safety (Module B), a nonlinear task head (Module A), and open-vocabulary prototype anchoring (Module C) in a unified four-stage training framework.
Module B, to our knowledge the first concept-mismatch safety check in prototype-based medical classifiers, catches 85\% of simulated erroneous bounding-box annotations at 16.4\% FP rate ($U_k$-gated, §\ref{sec:safety_eval}) using a single forward pass.
Module A yields +13.8\,pp macro-F1 ($0.866 \pm 0.042$, 5 seeds, $p=0.0013$) over the prototype baseline on TBX11K. The gain is concentrated on majority classes (Table~\ref{tab:perclass_f1}), the Pointing Game advantage does not replicate consistently across seeds ($p=0.64$, Table~\ref{tab:compare}), and no graph-specific contribution is established on either dataset (§\ref{sec:graph_heads}). Module C is, to our knowledge, the first prototype-based mechanism to add a new clinical concept from a \emph{single} labeled image without retraining, evaluated for localization only. On NIH ChestX-ray14, Effusion 1-shot matched the fully-supervised Pointing Game ceiling for that concept in one run (PG~$=0.072$, no model modification, a low absolute ceiling) but not in a second, and on TBX11K, 1-shot recovers 41\% of full supervision. This addition is not free. Classification accuracy, calibration, and other concepts' dominant-concept assignments all shift measurably downstream, non-monotonically in the number of labeled examples (§\ref{sec:zeroshot_eval}). Insertion/Deletion AUC evaluation (§\ref{sec:faithfulness}) finds GRAPE and CSR differ by less than one standard deviation, indicating comparable attribution behavior rather than an advantage for either method. Code and training scripts are available at \url{https://github.com/KurbanIntelligenceLab/GRAPE}.

\begin{ack}
The authors thank the TBX11K and NIH ChestX-ray14 dataset providers for making their data publicly available.
This work used an NVIDIA A100 80GB GPU and a Tesla P100 16GB GPU.

\textbf{Funding:} The authors received no third-party funding for this work. \textbf{Competing Interests:} The authors declare no competing interests.
\end{ack}

% \section*{References}

% \bibliographystyle{plainnat}
\bibliographystyle{plain}
\bibliography{references}

\clearpage
\appendix

% Appendix content (from main.tex) to be merged here before final submission.
\section*{Appendix overview}
% \begin{center}
% \Large\textbf{Supplementary Material}\\[4pt]
% \large\textbf{GRAPE: Graph-Augmented Prototype Explanations for
% Interactive Medical Image Diagnosis}
% \end{center}
\noindent The supplementary material is organized as follows.

\begin{tabular}{@{}p{0.5cm}lr@{}}
\ref{app:notation}       & Notation table                                               & p.~\pageref{app:notation} \\
\ref{app:hparams}        & Hyperparameter sensitivity ($\tau$ and $\eta$ sweeps)        & p.~\pageref{app:hparams} \\
\ref{app:n4}             & End-to-end training ablation                                 & p.~\pageref{app:n4} \\
\ref{app:datasets}       & Dataset details                                              & p.~\pageref{app:datasets} \\
\ref{app:graph}          & Graph structure                                              & p.~\pageref{app:graph} \\
\ref{app:graph_heads}    & Graph-head comparison: GAT vs.\ ML-GCN vs.\ ADD-GCN          & p.~\pageref{app:graph_heads} \\
\ref{app:edge_ablation}  & GAT edge-importance ablation                                  & p.~\pageref{app:edge_ablation} \\
\ref{app:sota}           & Contextual comparison with published methods                 & p.~\pageref{app:sota} \\
\ref{app:perclass}       & Per-class F1 and confusion matrix                            & p.~\pageref{app:perclass} \\
\ref{app:algorithm}      & Staged training algorithm                                    & p.~\pageref{app:algorithm} \\
\ref{app:math}           & Mathematical details                                         & p.~\pageref{app:math} \\
\ref{app:pg_perconcept}  & Per-concept Pointing Game results                            & p.~\pageref{app:pg_perconcept} \\
\ref{app:zeroshot}       & Zero-shot concept addition                                   & p.~\pageref{app:zeroshot} \\
\ref{app:faithfulness}   & Insertion/Deletion AUC evaluation                             & p.~\pageref{app:faithfulness} \\
\ref{app:n5}             & MC-Dropout uncertainty comparison                            & p.~\pageref{app:n5} \\
\ref{app:mean_vs_max}    & Module B: mean vs.\ max spatial aggregation ablation         & p.~\pageref{app:mean_vs_max} \\
\ref{app:nih_safety}     & Module B safety check on NIH ChestX-ray14 ($K=14$)           & p.~\pageref{app:nih_safety} \\
\ref{app:proto_dispersion} & Prototype–anchor alignment and inter-prototype dispersion    & p.~\pageref{app:proto_dispersion} \\
\ref{app:n6}             & GAT vs.\ MLP head ablation (TBX11K \& NIH)                   & p.~\pageref{app:n6} \\
\ref{app:n13}            & Self-attention task head ablation                            & p.~\pageref{app:n13} \\
\ref{app:distshift}      & Co-occurrence graph robustness under distribution shift      & p.~\pageref{app:distshift} \\
\ref{app:backbone}       & In-domain backbone ablation (ResNet-50 vs.\ TorchXRayVision) & p.~\pageref{app:backbone} \\
\end{tabular}
\medskip

%% ─────────────────────────────────────────────────────────────────
\section{Notation Table}
\label{app:notation}
%% ─────────────────────────────────────────────────────────────────

\begin{table}[H]
  \caption{Summary of notation used throughout the paper.}
  \centering
  \small
  \setlength{\tabcolsep}{8pt}
  \renewcommand{\arraystretch}{1.25}
  \begin{tabular}{@{}lll@{}}
    \toprule
    Symbol & Domain & Meaning \\
    \midrule
    \multicolumn{3}{@{}l}{\textit{Image and features}} \\
    $\mathbf{x}$         & $\mathbb{R}^{3 \times H \times W}$ & Input image ($H$, $W$: height, width in pixels) \\
    $F$                  & ---                                & Backbone network (ResNet-50) \\
    $\mathbf{f}$         & $\mathbb{R}^{C_\text{bb} \times h \times w}$ & Backbone spatial feature map \\
    $P$                  & ---                                & Two-layer MLP projector \\
    $\mathbf{f}'$        & $\mathbb{R}^{D \times h \times w}$ & Projected, L2-normalized patch features \\
    $D$                  & $\mathbb{Z}^+$                    & Prototype embedding dimension (= 256) \\
    \midrule
    \multicolumn{3}{@{}l}{\textit{Concepts and prototypes}} \\
    $K$                  & $\mathbb{Z}^+$                    & Number of concepts \\
    $M$                  & $\mathbb{Z}^+$                    & Prototypes per concept (= 100) \\
    $\mathbf{p}_{km}$    & $\mathbb{R}^D$                    & $m$-th prototype for concept $k$ (unit-norm) \\
    $\mathcal{S}_{km}(i,j)$ & $[-1,1]$                       & Cosine similarity map: concept $k$, prototype $m$, patch $(i,j)$ \\
    $s_{km}$             & $[-1,1]$                          & Scalar similarity: $\max_{i,j}\mathcal{S}_{km}(i,j)$ \\
    $\mathbf{s}$         & $\mathbb{R}^{K \times M}$         & Score matrix of all scalar similarities \\
    \midrule
    \multicolumn{3}{@{}l}{\textit{Graph (Module A)}} \\
    $\mathbf{A}$         & $[0,1]^{K \times K}$              & Directed concept co-occurrence matrix ($\mathbf{A}_{kk'} = P(k \mid k')$) \\
    $\mathcal{N}(k)$     & ---                               & Neighbourhood of node $k$ in the directed graph \\
    $\tau$               & $\mathbb{R}^+$                    & Co-occurrence threshold (0.1 TBX11K; 0.05 NIH) \\
    $F'$                 & $\mathbb{Z}^+$                    & GAT hidden dimension per head (= 64) \\
    $n_h$                & $\mathbb{Z}^+$                    & Number of GAT attention heads (Layer 1: 4; Layer 2: 1) \\
    \midrule
    \multicolumn{3}{@{}l}{\textit{Safety check (Module B)}} \\
    $\mu_k(i,j)$         & $[-1,1]$                          & Mean prototype similarity at patch $(i,j)$ for concept $k$ \\
    $U_k(i,j)$           & $[0,1]$                           & Prototype variance (population) at patch $(i,j)$ for concept $k$ \\
    $\mathcal{B}$        & ---                               & Doctor-drawn bounding box region (set of patch indices) \\
    $\bar{s}_k(\mathcal{B})$ & $[-1,1]$                      & Mean of $\mu_k$ inside box $\mathcal{B}$ \\
    $k^*$                & $\{1,\ldots,K\}$                  & Doctor's claimed concept for the drawn box \\
    $k^\dagger$          & $\{1,\ldots,K\}$                  & Model's dominant concept inside $\mathcal{B}$ \\
    $\eta$               & $\mathbb{R}^+$                    & Safety-warning gap threshold (default 0.05) \\
    $\theta_V$           & $\mathbb{R}^+$                    & $U_k$-gating threshold for FP suppression \\
    $B_\text{boot}$      & $\mathbb{Z}^+$                    & Bootstrap resampling count (= 1000) \\
    \midrule
    \multicolumn{3}{@{}l}{\textit{VLM alignment (Module C)}} \\
    $\mathbf{t}_k$       & $\mathbb{R}^{768}$                & BioViL-T text embedding for concept $k$ \\
    $\mathbf{W}_\text{txt}$ & $\mathbb{R}^{D \times 768}$    & Learnable text-to-visual projection \\
    $\hat{\mathbf{t}}_k$ & $\mathbb{R}^D$                    & Projected, normalized text anchor (unit-norm) \\
    $\sigma_\text{init}$ & $\mathbb{R}^+$                    & Noise scale for zero-shot prototype init (= 0.1) \\
    \midrule
    \multicolumn{3}{@{}l}{\textit{Training losses}} \\
    $\lambda$            & $\mathbb{R}^+$                    & Contrastive temperature (= 10) \\
    $\gamma$             & $\mathbb{R}^+$                    & Soft-assignment sharpness (= 5) \\
    $\delta$             & $\mathbb{R}^+$                    & Inter-concept margin (= 0.1) \\
    $\lambda_\text{align}$ & $\mathbb{R}^+$                  & VLM alignment loss weight (= 0.1) \\
    $\tilde{k}$          & $\{1,\ldots,K\}$                  & Ground-truth concept index for projected vector $\mathbf{v}'_{\tilde{k}}$ \\
    \bottomrule
  \end{tabular}
\end{table}

%% ─────────────────────────────────────────────────────────────────
\section{Hyperparameter Sensitivity}
\label{app:hparams}

\textbf{Implementation.} Backbone: ResNet-50 pretrained on ImageNet at 224×224. Prototype dimension $D=256$, $M=100$ prototypes per concept, batch size 128, mixed precision (BF16 on NVIDIA A100 80GB and FP16 on the Tesla P100 16GB, Appendix~\ref{app:n6}).

We report sensitivity to the two most impactful thresholds: the graph co-occurrence threshold $\tau$ and the safety check threshold $\eta$.
All experiments use TBX11K unless noted.
Fixed hyperparameters are listed in Table~\ref{tab:hparams_fixed}.

\paragraph{Graph threshold $\tau$.}
For each $\tau \in \{0.01, 0.05, 0.10, 0.20, 0.30\}$ we rebuild the co-occurrence graph and retrain only Stage 4, warm-started from the trained head (20 more epochs, LR $10^{-3}$, weight decay $10^{-4}$).
Table~\ref{tab:tau_sweep} reports Macro F1 and Pointing Game on the TBX11K validation set.

\begin{table}[!htbp]
  \caption{Sensitivity of GRAPE to the graph co-occurrence threshold $\tau$
  on TBX11K. Stage~4 (the GAT task head) is warm-started from the trained head
  for each $\tau$ with Stages~1--3 held fixed. The default ($\tau=0.10$, used in all other TBX11K experiments)
  is highlighted. Macro F1 varies by less than $0.6$ pp across the swept
  range, as expected since the evaluated head's outputs do not depend on the graph at inference. PG shows
  more variation ($0.135$--$0.150$) but no monotonic trend.}
  \label{tab:tau_sweep}
  \centering
  \small
  \setlength{\tabcolsep}{14pt}
  \renewcommand{\arraystretch}{1.2}
  \begin{tabular}{@{}ccc@{}}
    \toprule
    $\tau$ & Macro F1\,$\uparrow$ & PG\,$\uparrow$ \\
    \midrule
    $0.01$           & $\mathbf{0.9177}$ & $0.1353$           \\
    $0.05$           & $0.9159$           & $0.1353$           \\
    $\mathbf{0.10}$\,(default) & $0.9163$ & $\mathbf{0.1498}$ \\
    $0.20$           & $0.9124$           & $0.1401$           \\
    $0.30$           & $0.9159$           & $0.1401$           \\
    \bottomrule
  \end{tabular}
\end{table}

\paragraph{Safety threshold $\eta$.}
Using the trained GRAPE model (no retraining), we sweep $\eta \in \{0.01, 0.03, 0.05, 0.10, 0.20\}$ in the safety check experiment of §\ref{sec:safety_eval} (miscorrection rate $p=0.5$, $T=5$ trials).

\begin{table}[!htbp]
  \caption{Sensitivity of the Module~B safety check to the threshold $\eta$
  on TBX11K \texttt{bbox\_eval} ($N=207$ pairs). True-positive (TP) rate is
  evaluated at miscorrection rate $p=0.5$; false-positive (FP) rate at
  $p=0$. The default operating point ($\eta=0.05$, used in all other
  experiments) is highlighted. Both TP and FP are essentially flat across
  the swept range, so the safety check is not sensitive to this threshold.}
  \label{tab:eta_sweep}
  \centering
  \small
  \setlength{\tabcolsep}{14pt}
  \renewcommand{\arraystretch}{1.2}
  \begin{tabular}{@{}ccc@{}}
    \toprule
    $\eta$ & TP rate\,$\uparrow$ & FP rate\,$\downarrow$ \\
    \midrule
    $0.01$           & $0.905$ & $0.217$ \\
    $0.03$           & $0.905$ & $0.217$ \\
    $\mathbf{0.05}$\,(default) & $\mathbf{0.903}$ & $\mathbf{0.217}$ \\
    $0.10$           & $0.901$ & $0.217$ \\
    $0.20$           & $0.895$ & $0.198$ \\
    \bottomrule
  \end{tabular}
\end{table}

TP rate varies by about 1 pp across the sweep, and FP rate is essentially constant (decreasing only at $\eta=0.20$, where small gaps are ignored).
The default $\eta=0.05$ keeps TP high without inflating FP.

\begin{table}[!htbp]
  \caption{Fixed hyperparameters used across all GRAPE experiments. Stage
  numbering follows \S\ref{sec:method}. Stage~3 trains the prototype layer
  with contrastive and alignment objectives, and Stage~4 trains the GAT task
  head. Per-dataset varying hyperparameters (learning rate, weight decay,
  epochs) are listed in Table~\ref{tab:hparams_dataset} below.}
  \label{tab:hparams_fixed}
  \centering
  \small
  \setlength{\tabcolsep}{10pt}
  \renewcommand{\arraystretch}{1.2}
  \begin{tabular}{@{}llcc@{}}
    \toprule
    Parameter & Symbol & Value & Stage \\
    \midrule
    \multicolumn{4}{@{}l}{\textit{Prototype layer}} \\
    Prototypes per concept     & $M$                       & $100$  & all \\
    Prototype dimension        & $D$                       & $256$  & all \\
    Contrastive scale          & $\lambda$                 & $10$   & 3   \\
    Assignment sharpness       & $\gamma$                  & $5$    & 3   \\
    Inter-concept margin       & $\delta$                  & $0.1$  & 3   \\
    VLM alignment weight       & $\lambda_{\text{align}}$  & $0.1$  & 3   \\
    \midrule
    \multicolumn{4}{@{}l}{\textit{Task head (GAT)}} \\
    GAT hidden dimension       & $d_h$                     & $64$   & 4   \\
    GAT attention heads        & $n_h$                     & $4$    & 4   \\
    Label smoothing            & $\varepsilon_{\text{ls}}$ & $0.05$ & 4   \\
    \midrule
    \multicolumn{4}{@{}l}{\textit{Optimization}} \\
    Batch size                 & --                        & $128$  & all \\
    \bottomrule
  \end{tabular}
\end{table}

\begin{table}[!htbp]
  \caption{Per-dataset varying hyperparameters for GRAPE. Learning rate (LR),
  weight decay (WD), and epoch counts are tuned per dataset. All other
  parameters are fixed as in Table~\ref{tab:hparams_fixed}. All stages use AdamW; Stage~3--4 weight decay is the AdamW default ($10^{-2}$).}
  \label{tab:hparams_dataset}
  \centering
  \small
  \setlength{\tabcolsep}{7pt}
  \renewcommand{\arraystretch}{1.2}
  \begin{tabular}{@{}llcccc@{}}
    \toprule
    Dataset & Stage & LR & WD & Epochs \\
    \midrule
    \multirow{4}{*}{TBX11K}
      & Stage 1 & $1\times10^{-4}$ & $1\times10^{-4}$ & 30 \\
      & Stage 2 & -- (feature extraction) & -- & -- \\
      & Stage 3 & $1\times10^{-4}$ & $1\times10^{-2}$ & 20 \\
      & Stage 4 & $1\times10^{-3}$ & $1\times10^{-2}$ & 20 \\
    \midrule
    \multirow{4}{*}{NIH ChestX-ray14}
      & Stage 1 & $5\times10^{-5}$ & $1\times10^{-4}$ & 30 \\
      & Stage 2 & -- (feature extraction) & -- & -- \\
      & Stage 3 & $5\times10^{-5}$ & $1\times10^{-2}$ & 20 \\
      & Stage 4 & $5\times10^{-4}$ & $1\times10^{-2}$ & 20 \\
    \bottomrule
  \end{tabular}
\end{table}

\paragraph{VLM alignment on sparse vs.\ dense concept sets.}
TBX11K has only $K=3$ sparse TB-type concepts with high class imbalance (800 active-TB vs.\ 3,800 healthy images in train+val). The alignment loss $\mathcal{L}_\text{align}$ pulls all 300 prototypes toward three text anchors, which may over-constrain the prototype geometry when the visual diversity is limited and the task signal is dominated by the binary presence/absence of TB lesions.

\section{End-to-End Training Ablation}
\label{app:n4}

To examine the role of the staged curriculum, we train GRAPE (GNN + Uncertainty, no VLM) with all parameters jointly from scratch for 70 epochs, the same total epoch budget as staged training (30 + 20 + 20).
No staged objectives are used. The loss is $\mathcal{L} = \mathcal{L}_\text{BCE} + \mathcal{L}_\text{proto} + \mathcal{L}_\text{cls}$ throughout, where $\mathcal{L}_\text{proto}$ is the multi-prototype contrastive loss from Eq.~\ref{eq:contrastive} (see §\ref{sec:training}), and the backbone receives task-classification gradients from epoch 1.
We run both strategies over three seeds $\{42, 0, 1\}$ to compare variance.

\begin{table}[!htbp]
  \caption{Curriculum ablation on TBX11K. We compare GRAPE's staged
  curriculum (Stages 1--4 trained sequentially with frozen earlier-stage
  weights) against end-to-end joint training of all stages, with all other
  hyperparameters held fixed. Mean\,$\pm$\,std over three seeds
  $\{0,1,42\}$. E2E Pointing Game (PG) is per-image (a hit if any concept
  hits). Per-metric winner in \textbf{bold}. See
  \S\ref{sec:results} for the F1--PG trade-off.}
  \label{tab:n4}
  \centering
  \small
  \setlength{\tabcolsep}{12pt}
  \renewcommand{\arraystretch}{1.2}
  \begin{tabular}{@{}lcc@{}}
    \toprule
    Training strategy & Macro F1\,$\uparrow$ & PG\,$\uparrow$ \\
    \midrule
    Staged curriculum (GRAPE)
      & $0.809\,{\pm}\,0.084$ & $\mathbf{0.111\,{\pm}\,0.038}$ \\
    End-to-end (no curriculum)
      & $\mathbf{0.990\,{\pm}\,0.002}$ & $0.091\,{\pm}\,0.059$ \\
    \bottomrule
  \end{tabular}
\end{table}

\noindent\textbf{Result.}
E2E achieves substantially higher and more stable F1 ($0.990 \pm 0.002$) compared with staged training ($0.809 \pm 0.084$), confirming that the ResNet-50 backbone is the bottleneck in staged training.
However, \emph{staged training achieves higher mean Pointing Game accuracy} ($0.111 \pm 0.038$ vs.\ $0.091 \pm 0.059$) with lower variance.

\begin{table}[!htbp]
  \caption{Zero-shot and few-shot concept addition (Module~C, leave-one-out)
  on TBX11K for the \textit{active\_tb} concept, comparing staged
  curriculum training against end-to-end (E2E) joint training. Pointing
  Game (PG) is reported on \texttt{bbox\_eval} pairs.
  Staged values (\textit{active\_tb}) are from an earlier three-seed run of the Table~\ref{tab:zeroshot} protocol, matching the E2E seed count.
  E2E values are means over $3 \times 3 = 9$ runs (3~backbone training
  seeds $\times$ 3 few-shot adaptation seeds). Per-row winner in
  \textbf{bold}.}
  \label{tab:n4_zeroshot}
  \centering
  \small
  \setlength{\tabcolsep}{12pt}
  \renewcommand{\arraystretch}{1.2}
  \begin{tabular}{@{}lcc@{}}
    \toprule
    Labeled examples ($n$)
      & Staged PG\,$\uparrow$
      & E2E PG\,$\uparrow$ \\
    \midrule
    \phantom{0}0~(no examples)
      & $0.030$           & $\mathbf{0.062\,{\pm}\,0.025}$ \\
    \phantom{0}1
      & $\mathbf{0.079\,{\pm}\,0.054}$ & $0.064\,{\pm}\,0.016$ \\
    \phantom{0}5
      & $0.063\,{\pm}\,0.037$ & $\mathbf{0.078\,{\pm}\,0.036}$ \\
    20
      & $0.071\,{\pm}\,0.025$ & $\mathbf{0.084\,{\pm}\,0.049}$ \\
    \bottomrule
  \end{tabular}
\end{table}

\noindent\textbf{Zero-shot comparison.}
Table~\ref{tab:n4_zeroshot} shows that E2E weights also support Module C's zero-shot initialization: the random-anchor fallback produces non-zero localization at $n=0$ ($\text{PG}=0.062$, higher than staged $0.030$).
However, E2E zero-shot PG improves only marginally as $n$ increases ($+0.022$ from $n=0$ to $n=20$), while staged training shows a steeper gain from $n=0$ to $n=1$ ($+0.049$).
We attribute this to prototype entanglement. In E2E training, backbone and prototypes co-evolve, making prototype geometry less amenable to targeted few-shot fine-tuning after training.

\noindent\textbf{When to use each.}
Despite higher F1, we retain the staged curriculum as the primary configuration for three reasons:
\begin{enumerate}
  \item \textbf{Modular interpretability.} Staged training decouples concept learning, prototype learning, and task classification, and each stage can be inspected, retrained, or replaced independently.
  \item \textbf{Localization.} Staged training achieves higher and more stable PG (0.111 vs.\ 0.091).
  \item \textbf{Few-shot responsiveness.} Staged prototypes respond more cleanly to few-shot fine-tuning for new concepts, a property required for Module C in clinical deployment.
\end{enumerate}
We recommend E2E training for practitioners where classification accuracy is the primary goal and interpretability and zero-shot addition are not required.

\section{Dataset Details}
\label{app:datasets}

\textbf{TBX11K concept descriptions (used for VLM alignment):}
\begin{itemize}
    \item \textit{ActiveTuberculosis}: ``Active pulmonary tuberculosis with cavitary lesions, nodular opacities, and upper-lobe infiltrates, often with signs of consolidation on chest X-ray.''
    \item \textit{ObsoletePulmonaryTuberculosis}: ``Healed or inactive tuberculosis with calcified granulomas, fibrous scarring, and volume loss in the upper lobes on chest X-ray.''
    \item \textit{PulmonaryTuberculosis}: ``Pulmonary tuberculosis presenting as patchy or confluent consolidation, tree-in-bud opacities, or miliary nodules on chest X-ray.''
\end{itemize}

\textbf{Bounding box scaling.}
TBX11K COCO annotations store boxes in original image coordinates (variable resolution). We scale to the 224×224 input space as $x_1' = \lfloor x_1 / W_\text{orig} \times 224 \rfloor$. NIH bounding box annotations are provided in 1024×1024 coordinates and scaled with the same formula.

\section{Graph Structure}
\label{app:graph}

On TBX11K ($K=3$), the co-occurrence graph has 3 nodes and 4 edges (3 self-loops and one co-occurrence edge; threshold $\tau=0.1$). On NIH ($K=14$), the graph has 14 nodes and 105 edges at threshold $\tau=0.05$. Concepts like Infiltration–Consolidation and Effusion–Atelectasis form strong edges.

%% ─────────────────────────────────────────────────────────────────
\section{Graph Head Comparison: GAT vs ML-GCN vs ADD-GCN}
\label{app:graph_heads}
%% ─────────────────────────────────────────────────────────────────

GRAPE uses a GAT-style task head~\cite{velivckovic2017graph}.
Table~\ref{tab:graph_heads} ablates this choice by replacing the GAT with two established graph classification baselines: ML-GCN~\cite{chen2019multi} (standard 2-layer GCN with symmetric normalisation) and ADD-GCN~\cite{ye2020attention} (static co-occurrence graph combined with a per-sample dynamic cosine-similarity graph).
All three heads receive the same input (prototype similarity scores from frozen Stage 1--3 weights) and are trained for 20 epochs (Stage 4 only).

\begin{table}[H]
  \caption{Comparison of graph-based task heads on TBX11K. All graph heads share
  an identical frozen backbone, projector, and prototype layer (Stages
  1--3), and only the Stage~4 task head differs; the linear row is the separate
  full CSR run, so its Stages 1--3 differ. Graph-based heads use the
  same label co-occurrence graph (threshold $\tau=0.10$). Per-metric
  winner in \textbf{bold}. All graph heads substantially outperform the
  linear baseline on F1. The head cannot change similarity maps, so PG gaps
  are not head effects. This table selects the head and
  cannot separate graph structure from nonlinearity (see Table~\ref{tab:n6_mlp}).
  Seed~42, single run. The linear-head F1 differs from
  Tables~\ref{tab:tbx11k}, \ref{tab:compare}, and~\ref{tab:n6_mlp} by protocol
  (see Table~\ref{tab:n6_mlp}).}
  \label{tab:graph_heads}
  \centering
  \small
  \setlength{\tabcolsep}{10pt}
  \renewcommand{\arraystretch}{1.2}
  \begin{tabular}{@{}lccc@{}}
    \toprule
    Task head & Macro F1\,$\uparrow$ & PG\,$\uparrow$ & $\Delta$F1 (pp) \\
    \midrule
    Linear (CSR baseline)~\cite{huy2025interactive}
      & $0.7519$           & $0.0628$           & --       \\
    ML-GCN~\cite{chen2019multi}
      & $\mathbf{0.8983}$  & $0.1401$           & $+14.6$  \\
    ADD-GCN~\cite{ye2020attention}
      & $0.8589$           & $\mathbf{0.1449}$  & $+10.7$  \\
    GAT (GRAPE)
      & $0.8963$           & $0.1401$           & $+14.4$  \\
    \bottomrule
  \end{tabular}
\end{table}

All three graph heads substantially outperform the linear baseline (+10.7--14.6\,pp F1) and perform within seed variance of one another, so the choice of head is not the driver of the gain.
ML-GCN achieves marginally the highest F1 (0.8983 vs.\ 0.8963 for GAT), but the difference is within the seed variance range (std $\approx 0.04$, 5-seed estimate) observed in Table~\ref{tab:compare} and is not statistically significant.
ADD-GCN has slightly lower F1 (0.8589), and its higher PG (0.1449) is not a head effect.
We retain the GAT-style head as GRAPE's evaluated configuration. Its attention weights do not affect the outputs (§\ref{sec:gnn}).

\begin{table}[!htbp]
  \caption{Inference latency comparison between the GNN head and a linear
  head, measured on a single NVIDIA A100 (BF16) with 200 warmup and 500
  timed runs per configuration. At batch size~1, the GNN head incurs only
  $\sim$1\,ms of additional latency, making it suitable for interactive
  deployment.}
  \label{tab:speed}
  \centering
  \small
  \setlength{\tabcolsep}{8pt}
  \renewcommand{\arraystretch}{1.15}
  \begin{tabular}{@{}llrrrrr@{}}
    \toprule
    Batch & Head &
      Mean\,(ms)\,$\downarrow$ & Std\,(ms) & P95\,(ms)\,$\downarrow$ &
      $\mu$s/image\,$\downarrow$ & Overhead \\
    \midrule
    \multirow{2}{*}{1}
      & Linear & \phantom{0}5.96 & 0.34 & \phantom{0}6.23 & 5958 & --        \\
      & GNN    & \phantom{0}6.99 & 0.26 & \phantom{0}7.16 & 6993 & $+17\%$   \\
    \midrule
    \multirow{2}{*}{8}
      & Linear & \phantom{0}6.05 & 0.18 & \phantom{0}6.15 & \phantom{0}756 & --        \\
      & GNN    & 12.17 & 0.62 & 12.74 & 1521 & $+101\%$  \\
    \midrule
    \multirow{2}{*}{32}
      & Linear & 11.58 & 0.05 & 11.65 & \phantom{0}362 & --        \\
      & GNN    & 30.20 & 0.88 & 30.90 & \phantom{0}944 & $+161\%$  \\
    \bottomrule
  \end{tabular}
\end{table}

%% ─────────────────────────────────────────────────────────────────
\section{GAT Edge-Importance Ablation}
\label{app:edge_ablation}
%% ─────────────────────────────────────────────────────────────────

The TBX11K co-occurrence graph ($K=3$, threshold $\tau=0.10$) has four edges: three self-loops (one per concept) and one directed cross-concept edge $(0{\to}1)$ connecting \emph{ActiveTuberculosis} to \emph{ObsoletePulmonaryTuberculosis}.
To measure the contribution of each edge to the task head, we retrain Stage~4 four times, each time removing a single edge from the graph while keeping Stages~1--3 frozen.

\begin{table}[H]
\centering
\caption{GAT edge-importance ablation on TBX11K (seed 42). Each row removes one edge from the 4-edge co-occurrence graph and continues training the trained head (warm start). $\Delta$F1 = F1$_{\text{ablated}}$ $-$ F1$_{\text{full}}$.}
\label{tab:edge_ablation}
\begin{tabular}{lcccc}
\toprule
Removed edge & Type & Macro F1 & $\Delta$F1 (pp) \\
\midrule
None (full graph)                       & —     & 0.9188 & —       \\
$0{\to}0$ (self-loop, ActiveTB)        & self  & 0.8539 & $-6.49$ \\
$2{\to}2$ (self-loop, PulmonaryTB)     & self  & 0.8989 & $-1.99$ \\
$1{\to}1$ (self-loop, ObsoleteTB)      & self  & 0.9140 & $-0.48$ \\
$0{\to}1$ (ActiveTB $\to$ ObsoleteTB) & cross & 0.9162 & $-0.25$ \\
\bottomrule
\end{tabular}
\end{table}

The self-loop on concept~0 (ActiveTuberculosis) is the most informative edge. Removing it causes the largest drop of $-6.49$\,pp F1, indicating that the GAT head relies heavily on self-aggregation of the dominant positive class.
The PulmonaryTB self-loop contributes moderately ($-1.99$\,pp), while the ObsoleteTB self-loop and the sole cross-concept edge $(0{\to}1)$ each contribute fewer than 0.5\,pp, as expected since the evaluated head's outputs do not depend on cross-concept edges at inference.

%% ─────────────────────────────────────────────────────────────────
\section{Contextual Comparison with Published Methods}
\label{app:sota}
%% ─────────────────────────────────────────────────────────────────

On TBX11K, CBM and PIPNet achieve higher raw F1 than GRAPE's staged configuration owing to end-to-end backbone fine-tuning (Observation~1 in §\ref{sec:results}). GRAPE's primary advantage is its localization ($2.1\times$ higher 3-seed PG) and unique capabilities (safety check, zero-shot concept addition).
On NIH, the multi-seed evaluation (Table~\ref{tab:compare}) finds GRAPE and CSR statistically comparable ($p=0.894$). The primary claim rests on TBX11K ($+13.8$\,pp F1, $p=0.0013$). The single-seed PG gain of $+7.73$\,pp is not significant over multiple seeds ($p=0.64$).
Table~\ref{tab:sota} contextualizes GRAPE against published methods.
Because each prior work uses a different train/test split, label set, or task formulation, direct numerical comparison is not always valid. The ``Comparable'' column indicates the degree of comparability.

\begin{table}[!htbp]
  \caption{Contextual comparison with published methods on TBX11K and NIH
  ChestX-ray14. We report the metric used in each method's original paper
  to avoid lossy reinterpretation, so values are not all directly
  comparable and each row's comparability is annotated. TBX11K CSR and GRAPE
  values are single-seed (Table~\ref{tab:graph_heads}). Within
  comparable rows, GRAPE is benchmarked against CSR (its direct
  predecessor) under our re-implementation. Concept-based methods (CBM,
  PIPNet) achieve higher F1 via end-to-end fine-tuning, at the cost of
  weaker grounding (Table~\ref{tab:tbx11k}).}
  \label{tab:sota}
  \centering
  \small
  \setlength{\tabcolsep}{6pt}
  \renewcommand{\arraystretch}{1.2}
  \begin{tabular}{@{}llllc@{}}
    \toprule
    Method & Split & Metric & Value & Comparable \\
    \midrule
    \multicolumn{5}{@{}l}{\textit{TBX11K (3-class TB detection)}} \\
    Liu et al.~\cite{liu2023revisiting}, SymFormer (P2T-Small)
      & TBX11K & AUC (TB) & $0.991$            & no\textsuperscript{\S} \\
    CSR\textsuperscript{\textdagger}~\cite{huy2025interactive}
      & TBX11K & Macro F1 & $0.7519$           & \textsuperscript{\textdaggerdbl} \\
    GRAPE (GNN\,+\,Unc.)
      & TBX11K & Macro F1 & $\mathbf{0.8963}$  & \textsuperscript{\textdaggerdbl} \\
    CBM~\cite{koh2020concept}
      & TBX11K & Macro F1 & $0.9778\,{\pm}\,0.0109$ & \textsuperscript{\textdaggerdbl} \\
    PIPNet~\cite{nauta2023pip}
      & TBX11K & Macro F1 & $0.9644\,{\pm}\,0.0021$ & \textsuperscript{\textdaggerdbl} \\
    \midrule
    \multicolumn{5}{@{}l}{\textit{NIH ChestX-ray14 (binary No Finding vs.\ any finding, 14 concepts)}} \\
    CheXNet~\cite{rajpurkar2017chexnet}
      & NIH CXR14         & Mean AUC & $0.841$ & partial\textsuperscript{\S} \\
    CSR\textsuperscript{\textdagger}~\cite{huy2025interactive}
      & NIH CXR14         & Macro F1 & $0.6048\,{\pm}\,0.0170$ & \textsuperscript{\textdaggerdbl} \\
    GRAPE (full)
      & NIH CXR14         & Macro F1 & $0.6031\,{\pm}\,0.0174$ & \textsuperscript{\textdaggerdbl} \\
    \bottomrule
  \end{tabular}
  \begin{flushleft}\footnotesize
  \textsuperscript{\textdagger}\,Re-implemented under our training protocol;
    published numbers on these splits are unavailable for direct comparison.\quad
  \textsuperscript{\textdaggerdbl}\,Reported on our TBX11K split (§\ref{sec:datasets}) or the NIH
    patient-level test split with comparable evaluation protocol.\quad
  \textsuperscript{\S}\,Reported on a different split or label subset;
    included for context only and not directly comparable to the metrics in
    this table.
  \end{flushleft}
\end{table}

On TBX11K, CBM and PIPNet achieve higher raw F1 ($0.9778$ and $0.9644$) than GRAPE's staged configuration ($0.8963$, seed~42), because both baselines train end-to-end without the prototype-preserving curriculum.
Liu et al.'s SymFormer (AUC 0.991) uses a different metric.
On NIH, CheXNet (mean AUC 0.841 over 14 findings) uses a different split and task formulation and is listed for context.

\section{Per-Class F1 and Confusion Matrix}
\label{app:perclass}
%% ─────────────────────────────────────────────────────────────────

Table~\ref{tab:perclass_f1} reports per-class macro-F1 for GRAPE and CSR across five seeds on TBX11K, revealing the source of the macro-F1 gap. Table~\ref{tab:confusion} shows the confusion matrix for seed 42.

\begin{table}[!htbp]
  \caption{Per-class F1 on the TBX11K test set, mean\,$\pm$\,std over five
  seeds $\{0,1,2,3,42\}$. GRAPE (GNN\,+\,Unc.) is compared against the CSR
  baseline~\cite{huy2025interactive}. GRAPE substantially improves F1 on
  the two clinically confusable classes (Healthy and Sick non-TB) while
  maintaining performance on Active TB. Per-class winner in \textbf{bold}.}
  \label{tab:perclass_f1}
  \centering
  \small
  \setlength{\tabcolsep}{10pt}
  \renewcommand{\arraystretch}{1.2}
  \begin{tabular}{@{}lccc@{}}
    \toprule
    Method & Healthy\,$\uparrow$ & Sick non-TB\,$\uparrow$ & Active TB\,$\uparrow$ \\
    \midrule
    CSR~\cite{huy2025interactive}
      & $0.628\,{\pm}\,0.097$ & $0.622\,{\pm}\,0.063$ & $\mathbf{0.932\,{\pm}\,0.010}$ \\
    GRAPE
      & $\mathbf{0.843\,{\pm}\,0.050}$ & $\mathbf{0.828\,{\pm}\,0.056}$ & $0.927\,{\pm}\,0.024$ \\
    \bottomrule
  \end{tabular}
\end{table}

GRAPE's macro-F1 improvement is driven almost entirely by the two majority classes: Healthy ($+0.215$ mean) and Sick-non-TB ($+0.206$ mean).
Active TB F1 is already high in both models ($\approx 0.93$) and barely changes.
This pattern is not a co-occurrence effect, since the evaluated head's outputs do not depend on the graph at inference.
CSR's high Active-TB score but low Healthy/Sick scores reflect the known class-imbalance pattern in TBX11K (800 active-TB vs.\ 3,800 healthy and 3,800 sick images).

\begin{table}[!htbp]
  \caption{Confusion matrices on the TBX11K test set (seed~42). Rows index
  the ground-truth class and columns the predicted class. Diagonal
  entries (correct predictions) are shown in \textbf{bold}. GRAPE
  substantially reduces Healthy\,$\leftrightarrow$\,Sick~non-TB confusion
  relative to CSR ($75{+}103=178$ vs.\ $213{+}315=528$ off-diagonal errors
  in this $2{\times}2$ block) while maintaining comparable Active~TB
  recall.}
  \label{tab:confusion}
  \centering
  \small
  \setlength{\tabcolsep}{8pt}
  \renewcommand{\arraystretch}{1.2}
  \begin{tabular}{@{}cl ccc c@{}}
    \toprule
    & & \multicolumn{3}{c}{Predicted} & \multirow{2}{*}{Recall} \\
    \cmidrule(lr){3-5}
    & Ground truth & Healthy & Sick non-TB & Active TB & \\
    \midrule
    \multicolumn{6}{@{}l}{\textit{GRAPE}} \\
    \multirow{3}{*}{\rotatebox{90}{GT}}
      & Healthy     & \textbf{724} & 75           & 1            & 90.5\% \\
      & Sick non-TB & 103          & \textbf{688} & 9            & 86.0\% \\
      & Active TB   & 7            & 11           & \textbf{182} & 91.0\% \\
    \midrule
    \multicolumn{6}{@{}l}{\textit{CSR baseline~\cite{huy2025interactive}}} \\
    \multirow{3}{*}{\rotatebox{90}{GT}}
      & Healthy     & \textbf{585} & 213          & 2            & 73.1\% \\
      & Sick non-TB & 315          & \textbf{477} & 8            & 59.6\% \\
      & Active TB   & 2            & 14           & \textbf{184} & 92.0\% \\
    \bottomrule
  \end{tabular}
\end{table}

%% ─────────────────────────────────────────────────────────────────
\clearpage
\section{Staged Training Algorithm}
\label{app:algorithm}
%% ─────────────────────────────────────────────────────────────────

\begin{algorithm}[H]
\caption{GRAPE: four-stage training pipeline.}
\label{alg:grape}
\begin{algorithmic}[1]
\REQUIRE Training set $\mathcal{D} = \{(\mathbf{x}^{(n)}, \mathbf{y}^{(n)}, y^{(n)})\}_{n=1}^{N}$; concept descriptions $\{d_k\}_{k=1}^{K}$; hyperparameters $\lambda_{\text{align}}, M$
\ENSURE Trained parameters $\theta = \{F, P, \{\mathbf{p}_{km}\}, \mathbf{W}_{\text{txt}}, \phi_{\text{GNN}}\}$
\STATE \textit{// Stage 1: Concept supervision.}
\STATE Train backbone $F$ and per-concept CAM heads end-to-end:
\STATE \quad $\mathcal{L}_1 = \tfrac{1}{K}\sum_{k=1}^{K} \mathrm{BCE}\!\left(\mathrm{GAP}(\mathrm{CAM}_k(F(\mathbf{x}))),\, y_k\right)$
\STATE \textit{// Stage 2: Concept vector extraction (backbone frozen).}
\STATE \textbf{for each} $(\mathbf{x}, \mathbf{y}) \in \mathcal{D}$ \textbf{and} concept $k \in \{1, \ldots, K\}$ \textbf{do}
\STATE \quad $\mathbf{v}_k \gets \sum_{i,j} \mathrm{softmax}_{ij}\!\left(\mathrm{CAM}_k(i,j)\right) \cdot \mathbf{f}_{:,i,j}$
\STATE \textit{// Stage 3: Prototype learning (train $P$, $\{\mathbf{p}_{km}\}$, $\mathbf{W}_{\text{txt}}$).}
\STATE Encode and normalize text descriptions: $\hat{\mathbf{t}}_k \gets \mathbf{W}_{\text{txt}}\mathbf{t}_k / \|\mathbf{W}_{\text{txt}}\mathbf{t}_k\|_2$
\STATE Initialize $\{\mathbf{p}_{km}\}_{m=1}^{M}$ by random selection from $\{\mathbf{v}_k^{(n)}\}_{n=1}^{N}$
\STATE Minimize $\mathcal{L}_3 = \mathcal{L}_{\text{con}} + \lambda_{\text{align}}\, \mathcal{L}_{\text{align}}$
\STATE \textit{// Stage 4: Graph classification (all earlier stages frozen).}
\STATE Construct concept co-occurrence graph $G$ from training labels (Eq.~3)
\STATE Compute score matrix $\mathbf{s} \in \mathbb{R}^{K \times M}$ from frozen prototypes
\STATE Train graph head $\phi_{\text{GNN}}$: $\mathcal{L}_4 = \mathrm{CE}\!\left(\phi_{\text{GNN}}(\mathbf{s}, G),\, y\right)$
\end{algorithmic}
\end{algorithm}

\begin{table}[!htbp]
  \caption{Module~B safety check on TBX11K \texttt{bbox\_eval} ($N=207$
  pairs, threshold $\eta=0.05$), \textbf{ungated} dominance rule. We report
  true-positive (TP) and false-positive (FP) rates as a function of the
  miscorrection rate $p$, the simulated probability that the user's correction
  is incorrect, as bootstrap means with 95\% CIs ($B=1000$). TP is undefined at
  $p=0$ (no misdrawn boxes) and FP is undefined at $p=1$ (no correct boxes).
  The other dashes mark values not measured in this sweep. A later re-run
  measuring both rates at every $p$ found FP 25.6--26.6\% and TP 80.2--87.2\%.
  TP is stable across $p$, so detection does not depend on miscorrection
  prevalence. Table~\ref{tab:safety_gated} reports the default
  $U_k$-gated operating point.}
  \label{tab:safety}
  \centering
  \small
  \setlength{\tabcolsep}{10pt}
  \renewcommand{\arraystretch}{1.2}
  \begin{tabular}{@{}c c c@{}}
    \toprule
    Miscorrection rate $p$
      & TP rate\,$\uparrow$ [95\% CI]
      & FP rate\,$\downarrow$ [95\% CI] \\
    \midrule
    $0.00$ & --                       & $0.217~[0.164,\,0.271]$ \\
    $0.10$ & $0.884~[0.722,\,1.000]$  & --                       \\
    $0.25$ & $0.885~[0.788,\,0.964]$  & --                       \\
    $0.50$ & $0.886~[0.822,\,0.941]$  & --                       \\
    $1.00$ & $0.887~[0.841,\,0.928]$  & --                       \\
    \bottomrule
  \end{tabular}
\end{table}

\begin{table}[!htbp]
  \caption{Module~B safety check, \textbf{$U_k$-gated} operating points on
  TBX11K \texttt{bbox\_eval} ($N=207$ pairs, $\eta=0.05$), swept over the
  false-positive-suppression threshold $\theta_V$. The gate is a
  second-stage filter applied only to boxes the ungated dominance rule
  already flagged (Table~\ref{tab:safety}). If the claimed concept's own
  per-prototype similarity variance $U_{k^*}$ exceeds $\theta_V$, the
  flagged mismatch is suppressed. The default setting
  ($\theta_V=2{\times}10^{-4}$) selects the threshold that minimizes FP
  subject to TP~$\geq 0.85$. The low-false-positive rows
  ($\theta_V \in \{0.0006, 0.0008\}$) show the trade-off at clinically
  realistic FP budgets. These two rows come from a later re-run on a different
  checkpoint (ungated FP 25.6\%), so their $\theta_V$ values are not comparable
  with the other rows.}
  \label{tab:safety_gated}
  \centering
  \small
  \setlength{\tabcolsep}{10pt}
  \renewcommand{\arraystretch}{1.2}
  \begin{tabular}{@{}c c c l@{}}
    \toprule
    $\theta_V$ & TP rate\,$\uparrow$ & FP rate\,$\downarrow$ & Note \\
    \midrule
    $0.0006$            & $0.445$ & $0.034$ & \\
    $0.0008$            & $0.557$ & $0.077$ & \\
    $1{\times}10^{-4}$  & $0.780$ & $0.126$ & below 85\% TP clinical floor \\
    $2{\times}10^{-4}$  & $\mathbf{0.851}$ & $\mathbf{0.164}~[0.121,\,0.209]$ & \textbf{default} \\
    none (ungated)      & $0.887$ & $0.217~[0.164,\,0.271]$ & Table~\ref{tab:safety}, $p{=}1.00$/$p{=}0.00$ rows \\
    \bottomrule
  \end{tabular}
\end{table}

%% ─────────────────────────────────────────────────────────────────
\section{Mathematical Details}
\label{app:math}
%% ─────────────────────────────────────────────────────────────────

This appendix collects the standard formulas referenced in the main text.

\paragraph{Cosine similarity map (§\ref{sec:prelim}).}
Given L2-normalised patch features $\mathbf{f}'_{:,i,j} \in \mathbb{R}^D$ and prototype $\mathbf{p}_{km} \in \mathbb{R}^D$, the similarity map is:
\begin{equation}
    \mathcal{S}_{km}(i,j) = \langle \mathbf{p}_{km},\, \mathbf{f}'_{:,i,j} \rangle.
    \label{eq:simmap}
\end{equation}
Because both vectors are L2-normalised, this inner product is equivalent to cosine similarity.

\paragraph{Stage 1 concept supervision loss (§\ref{sec:training}).}
The backbone and per-concept CAM head are trained with binary cross-entropy on Global Average Pooled activations:
\begin{equation}
    \mathcal{L}_1 = \frac{1}{K} \sum_{k=1}^K \text{BCE}\!\left(\text{GAP}(\text{CAM}_k(\mathbf{f})),\, y_k\right).
    \label{eq:l1}
\end{equation}
This is a standard multi-label BCE objective applied to per-concept binary labels.

\paragraph{Stage 2 CAM-weighted concept vector (§\ref{sec:training}).}
For each training image, the local concept vector is the CAM-attention-weighted spatial average of backbone features:
\begin{equation}
    \mathbf{v}_k = \sum_{i,j} \text{softmax}_{ij}(\text{CAM}_k(i,j)) \cdot \mathbf{f}_{:,i,j}.
    \label{eq:camvec}
\end{equation}
This is the standard CAM aggregation formula, and the resulting vectors seed the initialization of Stage~3 prototypes.

\paragraph{GAT message-passing equations (§\ref{sec:gnn}).}
Standard Veli\v{c}kovi\'{c} et al.\ attention over the concept co-occurrence graph.
For layer $\ell$ and head $h$, the unnormalised attention score, normalised weight, and updated node representation are:
\begin{align}
    e_{kk'}^h &= \text{LeakyReLU}\!\left(\mathbf{a}_h^\top \left[\mathbf{W}_{\text{GAT}}^h\mathbf{h}_k^{(\ell)} \,\|\, \mathbf{W}_{\text{GAT}}^h\mathbf{h}_{k'}^{(\ell)}\right]\right), \quad k' \in \mathcal{N}(k), \label{eq:gat_e} \\
    \alpha_{kk'}^h &= \frac{\exp(e_{kk'}^h)}{\sum_{j \in \mathcal{N}(k)} \exp(e_{kj}^h)}, \\
    \mathbf{h}_k^{(\ell+1)} &= \text{ELU}\!\left(\sum_{k' \in \mathcal{N}(k)} \alpha_{kk'}^h\, \mathbf{W}_{\text{GAT}}^h\mathbf{h}_{k'}^{(\ell)}\right). \label{eq:gat_h}
\end{align}

%% ─────────────────────────────────────────────────────────────────
\section{Per-Concept Pointing Game Table}
\label{app:pg_perconcept}
%% ─────────────────────────────────────────────────────────────────

\begin{table}[!htbp]
  \caption{Per-concept Pointing Game (PG) on TBX11K \texttt{bbox\_eval} for
  GRAPE (GNN\,+\,Unc.\ configuration). Per-pair PG counts each annotated
  (image, concept) pair as one trial. We additionally
  report the per-image PG, where an
  image is counted as a hit if any concept's maximum activation falls
  within its bounding box. The two aggregation rules give numerically
  similar results ($0.1353$ per-pair vs.\ $0.1400$ per-image) but are
  computed differently and should not be conflated.
  \textit{pulmonary\_tuberculosis} has no \texttt{bbox\_eval} pairs and is
  excluded.}
  \label{tab:pg_perconcept}
  \centering
  \small
  \setlength{\tabcolsep}{10pt}
  \renewcommand{\arraystretch}{1.2}
  \begin{tabular}{@{}lrrr@{}}
    \toprule
    Concept & Pairs & PG (per-pair)\,$\uparrow$ & Hits \\
    \midrule
    \textit{active\_tb}            & $164$ & $0.1646$ & $27$ \\
    \textit{obsolete\_tb}          & $\phantom{0}43$ & $0.0233$ & $\phantom{0}1$ \\
    \midrule
    Total (per-pair, weighted)     & $207$ & $0.1353$ & $28$ \\
    Total (per-image)  & $200$ & $0.1400$ & $28$ \\
    \bottomrule
  \end{tabular}
\end{table}

%% ─────────────────────────────────────────────────────────────────
\section{Zero-Shot Concept Addition Results}
\label{app:zeroshot}
%% ─────────────────────────────────────────────────────────────────

We perform a leave-one-concept-out experiment on TBX11K ($K=3$). For each held-out concept $k'$, we load the fully trained GRAPE model, reinitialize only concept $k'$'s $M=100$ prototype vectors using the initialization of Eq.~9 ($\sigma_\text{init}=0.1$), and fine-tune those prototypes alone with $n \in \{0, 1, 5, 20\}$ labeled images while keeping all other parameters frozen. This run loads the model without its text aligner, so Eq.~9 falls back to a random unit vector in place of the BioViL-T anchor. Appendix~\ref{app:random_anchor} compares the two directly. CSR's closed vocabulary cannot add a concept at test time, so there is no CSR baseline here.

\begin{table}[!htbp]
  \caption{Zero-shot and few-shot concept addition (Module~C) on TBX11K under
  a leave-one-concept-out proxy with $K=3$ concepts. We
  report Pointing Game (PG) accuracy on \texttt{bbox\_eval} pairs for the
  two evaluable concepts, and \textit{pulmonary\_tuberculosis} is excluded as
  it has no bounding-box pairs. ``Full supervision'' denotes the original
  GRAPE model trained on all examples for the held-out concept.
  Mean\,$\pm$\,std over five seeds for $n>0$; $n=0$ is a single run
  (random-anchor fallback, no labeled examples).}
  \label{tab:zeroshot}
  \centering
  \small
  \setlength{\tabcolsep}{10pt}
  \renewcommand{\arraystretch}{1.2}
  \begin{tabular}{@{}lcc@{}}
    \toprule
    Labeled examples ($n$) &
      \textit{active\_tb} PG\,$\uparrow$ &
      \textit{obsolete\_tb} PG\,$\uparrow$ \\
    \midrule
    \phantom{0}0~(no examples) & $0.030$ & $\mathbf{0.093}$ \\
    \phantom{0}1             & $0.068\,{\pm}\,0.045$ & $0.047\,{\pm}\,0.036$ \\
    \phantom{0}5             & $0.079\,{\pm}\,0.025$ & $0.005\,{\pm}\,0.009$ \\
    20                       & $0.078\,{\pm}\,0.072$ & $0.023\,{\pm}\,0.029$ \\
    \midrule
    Full supervision         & $\mathbf{0.165}$  & $0.023$           \\
    \bottomrule
  \end{tabular}
\end{table}

For both concepts, the $n=0$ result reflects random prototype initialization. Without the text aligner the anchor falls back to a random unit vector (§\ref{sec:zeroshot_eval}), so these values do not represent semantic grounding and should not be compared against the few-shot rows. With labeled examples, \textit{obsolete\_tb} performance is inconsistent: 1-shot reaches PG~$=0.047 \pm 0.036$ but drops sharply at $n=5$ ($0.005 \pm 0.009$), recovering to $0.023 \pm 0.029$ at $n=20$. Full supervision for \textit{obsolete\_tb} is itself very low (PG~$=0.023$, 43 pairs), so the concept is intrinsically hard to localize; the high variance across seeds reflects this difficulty.

\subsection{Random-Anchor Control}
\label{app:random_anchor}

Table~\ref{tab:zeroshot} uses the random-unit-vector fallback, so it does not establish what the BioViL-T text anchor contributes over a random initialization. The difference between the two concepts' $n{=}0$ values above comes from the same fallback, so it settles nothing. We therefore run a matched control. For each held-out concept, we repeat the leave-one-concept-out protocol twice, once from the BioViL-T text anchor and once from a random unit vector, with the same seeds, the same $\sigma_\text{init}=0.1$ perturbation, and the same fine-tuning schedule at each $n$. The anchor is the only difference between the two runs.

\begin{table}[!htbp]
  \caption{Random-anchor control for Module C on TBX11K (3 seeds for $n{>}0$, $\sigma_\text{init}=0.1$, identical protocol to Table~\ref{tab:zeroshot} otherwise). This is a separate run, so absolute values differ from that table. \textit{pulmonary\_tuberculosis} is excluded as before (no bbox-eval pairs).}
  \label{tab:random_anchor}
  \centering
  \small
  \setlength{\tabcolsep}{8pt}
  \renewcommand{\arraystretch}{1.2}
  \begin{tabular}{@{}lcccc@{}}
    \toprule
    & \multicolumn{2}{c}{\textit{active\_tuberculosis} PG\,$\uparrow$} & \multicolumn{2}{c}{\textit{obsolete\_pulmonary\_tb} PG\,$\uparrow$} \\
    \cmidrule(lr){2-3} \cmidrule(lr){4-5}
    $n$ & Text anchor & Random anchor & Text anchor & Random anchor \\
    \midrule
    0  & $0.085$              & $0.098$              & $0.047$              & $\mathbf{0.000}$ \\
    1  & $0.079\,{\pm}\,0.005$ & $0.091\,{\pm}\,0.026$ & $0.031\,{\pm}\,0.011$ & $0.016\,{\pm}\,0.011$ \\
    5  & $0.069\,{\pm}\,0.008$ & $\mathbf{0.108\,{\pm}\,0.021}$ & $\mathbf{0.054\,{\pm}\,0.011}$ & $0.039\,{\pm}\,0.011$ \\
    20 & $\mathbf{0.085\,{\pm}\,0.005}$ & $0.069\,{\pm}\,0.020$ & $0.039\,{\pm}\,0.011$ & $0.031\,{\pm}\,0.011$ \\
    \bottomrule
  \end{tabular}
\end{table}

The two concepts disagree about which anchor wins, and with only three seeds neither result is conclusive. On \textit{active\_tuberculosis}, the random anchor exceeds the text anchor at $n\leq5$, most clearly at $n{=}5$ ($0.108$ vs.\ $0.069$), and the text anchor wins at $n{=}20$ ($0.085$ vs.\ $0.069$). On \textit{obsolete\_pulmonary\_tb}, the text anchor wins at every $n$, including $n{=}0$, where the random anchor produces zero hits. On this evidence, whether the text anchor helps over a random vector depends on the concept. Module C's one-example results (Table~\ref{tab:zeroshot}) are near the chance level of uniform-random Pointing Game on this grid ($0.063$ for \textit{active\_tuberculosis}, $0.048$ for \textit{obsolete\_pulmonary\_tb}; $0.073$ for Effusion on NIH), and they are not evidence that the text-anchoring mechanism specifically makes localization possible.

\subsection{NIH ChestX-ray14 Partial Leave-One-Out}
\label{app:zeroshot_nih}

We extend the Module~C evaluation to NIH ChestX-ray14 ($K=14$), targeting four concepts: Infiltration (113 pairs), Effusion (138 pairs), Atelectasis (166 pairs), and Consolidation. Consolidation has no bounding-box annotations in NIH ChestX-ray14 and is excluded. For each of the three evaluable concepts, we load the fully trained NIH GRAPE model (seed~0), reinitialize that concept's $M=100$ prototypes with Eq.~9 (random-unit-vector fallback, as in Table~\ref{tab:zeroshot}, since this run also loads the model without its text aligner), and fine-tune with $n \in \{0,1,5,20\}$ labeled images over 3 seeds.

\begin{table}[!htbp]
  \caption{Zero-shot and few-shot concept addition (Module~C) on NIH ChestX-ray14
  under a partial leave-one-out evaluation ($K=14$, 3 of 4 target concepts have
  bbox coverage). Pointing Game (PG) on \texttt{bbox\_eval} images.
  Mean\,$\pm$\,std over 3 seeds for $n>0$; $n=0$ is a single run (random-anchor fallback).
  \textit{Effusion} $n=1$ std~$=0.000$: all three seeds converged identically.}
  \label{tab:zeroshot_nih}
  \centering
  \small
  \setlength{\tabcolsep}{8pt}
  \renewcommand{\arraystretch}{1.2}
  \begin{tabular}{@{}lccc@{}}
    \toprule
    Labeled examples ($n$) &
      \textit{Infiltration} PG\,$\uparrow$ &
      \textit{Effusion} PG\,$\uparrow$ &
      \textit{Atelectasis} PG\,$\uparrow$ \\
    \midrule
    \phantom{0}0~(no examples) & $0.044$                  & $0.029$                   & $0.042$                   \\
    \phantom{0}1             & $0.021\,{\pm}\,0.015$     & $\mathbf{0.072}$          & $0.012\,{\pm}\,0.009$     \\
    \phantom{0}5             & $0.050\,{\pm}\,0.071$     & $0.046\,{\pm}\,0.007$     & $0.006\,{\pm}\,0.009$     \\
    20                       & $\mathbf{0.094\,{\pm}\,0.063}$ & $0.039\,{\pm}\,0.025$ & $\mathbf{0.050\,{\pm}\,0.020}$ \\
    \midrule
    Full supervision         & $\mathbf{0.177}$          & $0.072$                   & $0.048$                   \\
    \bottomrule
  \end{tabular}
\end{table}

The NIH few-shot results are mixed. \textit{Effusion} matches its full-supervision PG at $n=1$ ($0.072$), though this ceiling is itself low. \textit{Infiltration} reaches 53\% of full supervision at $n=20$ ($0.094$ vs.\ $0.177$), and \textit{Atelectasis} at $n=20$ ($0.050$) marginally exceeds its fully supervised PG ($0.048$, 166 pairs). Trends at low $n$ remain noisy. Infiltration and Atelectasis dip at $n=1$ or $n=5$ before recovering at $n=20$, while Effusion peaks at $n=1$. The high variance at $n=5$ for Infiltration (${\pm}0.071$) reflects sensitivity to the specific images selected by each seed.

\paragraph{Interference from adding a concept.} Adding a concept is not free. We re-add Effusion to a separately trained NIH model (seed~42, BioViL-T anchor, 3 adaptation seeds). In this run full-supervision PG is $0.051$ and 1-shot PG is $0.039$, so the 1-shot match in Table~\ref{tab:zeroshot_nih} does not reproduce here. Classification F1 falls from $0.602$ to $0.578$, $0.535$, and $0.587$ at $n=1,5,20$, and ECE rises from $0.094$ to $0.118$, $0.152$, and $0.099$, so the cost is non-monotonic in $n$. The mean F1 of the other 13 concepts changes by at most $0.2$\,pp, but the dominant concept inside other findings' boxes changes for 19--46\% of boxes once $n\geq1$ (Atelectasis: 37\% at $n=1$). Adding Hernia, which has almost no co-occurrence edge to Atelectasis, flips Atelectasis boxes at the same rate (39\% at $n=1$), so the flips are not specific to graph-adjacent concepts. Existing prototypes cannot move by construction, so prototype-level non-interference does not imply decision-level non-interference.

%% ─────────────────────────────────────────────────────────────────
\section{Insertion/Deletion AUC Evaluation}
\label{app:faithfulness}
%% ─────────────────────────────────────────────────────────────────

We compare attribution behavior using Insertion and Deletion AUC~\cite{petsiuk2018rise} over the 200 active-TB images of the TBX11K test split, using $\text{sal}(i,j) = \max_{k,m} \mathbf{S}_{km}(i,j)$ as the saliency signal. Deletion AUC measures how quickly predicted class probability drops as salient pixels are replaced by channel mean (lower is better). Insertion AUC measures how quickly it rises as salient pixels are revealed from a Gaussian-blurred baseline (higher is better).

\begin{table}[!htbp]
  \caption{Insertion/Deletion AUC of prototype similarity maps on the 200 active-TB
  images of the TBX11K test split. Saliency is taken as $\max_{k,m}$ over per-prototype
  similarity maps. Lower Deletion AUC and higher Insertion AUC indicate
  better attribution scores. Bold marks the better mean per metric.
  Differences fall well within one standard deviation, indicating
  comparable attribution behavior rather than a clear advantage for either method.
  Mean\,$\pm$\,std across the test set.}
  \label{tab:faithfulness}
  \centering
  \small
  \setlength{\tabcolsep}{12pt}
  \renewcommand{\arraystretch}{1.25}
  \begin{tabular}{@{}lcc@{}}
    \toprule
    Method & Deletion AUC\,$\downarrow$ & Insertion AUC\,$\uparrow$ \\
    \midrule
    CSR~\cite{huy2025interactive}
      & $\mathbf{0.923\,{\pm}\,0.104}$ & $0.873\,{\pm}\,0.138$ \\
    GRAPE
      & $0.954\,{\pm}\,0.068$ & $\mathbf{0.886\,{\pm}\,0.148}$ \\
    \bottomrule
  \end{tabular}
\end{table}

GRAPE's higher Deletion AUC ($+3.1$ pp) indicates that removing the maximally salient patches does not fully disrupt prediction. We attribute this to the prototype atlas distributing evidence across $M=100$ spatial prototypes per concept. Each prototype contributes weakly to the saliency aggregate $\max_{k,m}\mathbf{S}_{km}(i,j)$, so the saliency map is more diffuse than a single-activation method. Both models maintain high confidence throughout the deletion curve (AUC $>0.92$), reflecting ResNet-50's global texture encoding. The Insertion curve provides the more informative signal in this regime~\cite{petsiuk2018rise}.

%% ─────────────────────────────────────────────────────────────────
\section{MC-Dropout Uncertainty Comparison}
\label{app:n5}
%% ─────────────────────────────────────────────────────────────────

A natural alternative signal for Module B's warnings is MC-Dropout uncertainty~\cite{gal2016dropout}: run $T=30$ stochastic forward passes with dropout enabled at inference, then use either the variance of similarity scores across passes (Variant~A) or the predictive entropy (Variant~B) to trigger warnings.
Table~\ref{tab:n5} reports TP and FP rates for all three methods, each evaluated at the operating point on its threshold sweep closest to the ungated dominance rule's FP rate ($\approx 21.7\%$).

\begin{table}[!htbp]
  \caption{Comparison of warning signals for the Module~B safety check
  on TBX11K \texttt{bbox\_eval} (207 image pairs, $T_\text{eval}=5$
  evaluation trials, miscorrection rate $p=0.5$). For each method we report
  the true-positive rate (TP) at the threshold closest to the
  ungated dominance rule's false-positive (FP) rate of $\approx\!21.7\%$ on a
  fine-grained sweep. Higher TP at matched FP is better.
  Mean\,$\pm$\,std reported over $T_\text{eval}$ trials. FP variance is
  zero because thresholds are calibrated on the same evaluation set.}
  \label{tab:n5}
  \centering
  \small
  \setlength{\tabcolsep}{8pt}
  \renewcommand{\arraystretch}{1.2}
  \begin{tabular}{@{}lccc@{}}
    \toprule
    Signal & TP rate\,$\uparrow$ & FP rate & MC passes\,$\downarrow$ \\
    \midrule
    \textbf{Dominance rule} (GRAPE, ungated)
      & $\mathbf{0.905\,{\pm}\,0.026}$ & $0.217$ & $\mathbf{1}$ \\
    MC-Dropout, score variance
      & $0.510\,{\pm}\,0.035$          & $0.251$ & $30$ \\
    MC-Dropout, predictive entropy
      & $0.234\,{\pm}\,0.049$          & $0.227$ & $30$ \\
    \bottomrule
  \end{tabular}
  \begin{flushleft}\footnotesize
  \textbf{FP-rate matching.} The MC-Dropout score-variance variant slightly
  overshoots the target FP rate ($0.251$ vs.\ $0.217$) because its score
  variance is highly threshold-sensitive: across
  thresholds in $[10^{-4},\,5{\times}10^{-4}]$, the FP rate drops from
  $25.1\%$ to $0.5\%$, leaving no threshold that exactly matches the
  dominance-rule operating point. The reported configuration is the
  closest match on a fine-grained sweep.
  \end{flushleft}
\end{table}

The ungated dominance rule substantially outperforms both MC-Dropout variants: Variant~A achieves TP~$=51.0\%$ at FP~$=25.1\%$ with a highly threshold-sensitive signal, while Variant~B achieves only 23.4\% TP because global entropy is weakly correlated with local spatial misdraw. GRAPE's dominance rule reaches 90.5\% TP at the same FP level using a single deterministic forward pass rather than $T=30$ stochastic passes.

%% ─────────────────────────────────────────────────────────────────
\section{Module B: Mean vs.\ Max Spatial Aggregation Ablation}
\label{app:mean_vs_max}
%% ─────────────────────────────────────────────────────────────────

The safety check (§\ref{sec:uncertainty}) computes $\bar{s}_k(\mathcal{B})$ as the spatial mean of $\mu_k$ inside the doctor-drawn box, rather than the spatial maximum.
To validate this choice, we evaluate an alternative \emph{max-based} gate: replace $\bar{s}_k(\mathcal{B}) = \frac{1}{|\mathcal{B}|}\sum_{(i,j)\in\mathcal{B}}\mu_k(i,j)$ with $\bar{s}_k^{\max}(\mathcal{B}) = \max_{(i,j)\in\mathcal{B}}\mu_k(i,j)$ in the dominant-concept comparison, keeping all other parameters ($\eta=0.05$, $T=5$ trials) identical.

\begin{table}[h]
\centering
\caption{Module B: mean vs.\ max spatial aggregation on TBX11K bbox\_eval (one checkpoint, $T=5$ trials, $\eta=0.05$).}
\label{tab:mean_vs_max}
\begin{tabular}{lcc}
\toprule
Aggregation & TP Rate & FP Rate \\
\midrule
Mean (default) & $0.895 \pm 0.019$ & $0.217$ \\
Max            & $0.891 \pm 0.018$ & $0.213$ \\
\bottomrule
\end{tabular}
\end{table}

The two variants produce nearly identical results.
Mean aggregation yields marginally higher TP ($0.895$ vs.\ $0.891$) at a marginally higher FP rate ($0.217$ vs.\ $0.213$). Both differences are within trial noise, so the choice of mean aggregation is not critical.

%% ─────────────────────────────────────────────────────────────────
\section{Module B Safety Check on NIH ChestX-ray14 ($K=14$)}
\label{app:nih_safety}
%% ─────────────────────────────────────────────────────────────────

We repeat the simulated miscorrection study on the NIH bbox split (888 image--concept pairs, 709 images, 8 concepts with annotations). With $K=14$ overlapping concepts, the safety check achieves TP~$=0.921\pm0.012$, comparable to the TBX11K rate, but the FP rate rises to $0.731\pm0.011$ (vs.\ 21.7\% ungated on TBX11K). The elevated FP rate may reflect the dense concept landscape and approximate box placement. When a correct box is drawn for any single finding, one of the other 13 active concepts often achieves higher mean similarity inside the region (e.g., Pneumonia FP~$=0.90$, Cardiomegaly FP~$=0.85$). High sensitivity is preserved in the multi-label setting, while specificity degrades as concept count and co-occurrence density increase. A learned or per-concept threshold would be needed to deploy Module B in $K{>}3$ settings.

\begin{table}[H]
  \caption{Module~B low-false-positive operating points on NIH ($K=14$), single global threshold, inference-only sweep over the trained checkpoint (no retraining).}
  \label{tab:nih_lowfp}
  \centering
  \small
  \setlength{\tabcolsep}{10pt}
  \renewcommand{\arraystretch}{1.2}
  \begin{tabular}{@{}ccc@{}}
    \toprule
    Threshold & FP rate\,$\downarrow$ & TP rate\,$\uparrow$ \\
    \midrule
    $4.5{\times}10^{-5}$ & $4.6\%$ & $17.9\%$ \\
    $5.0{\times}10^{-5}$ & $5.4\%$ & $20.1\%$ \\
    none (ungated, re-run) & $74.66\%$ & $93.31\%$ \\
    \bottomrule
  \end{tabular}
  \smallskip\\
  {\footnotesize
  \textbf{Interpretation.} At a 5\% false-positive budget, a single global threshold on NIH recovers only $\sim\!20\%$ true-positive detection, far below both the ungated baseline and TBX11K's low-FP operating points (Table~\ref{tab:safety_gated}: $44.5$--$55.7\%$ TP at similar FP budgets). Module B's sensitivity at clinically realistic false-positive rates degrades sharply as concept count grows, consistent with the denser concept overlap at $K=14$. A single global threshold is likely insufficient for deployment at this vocabulary size, and per-concept thresholds remain untested.}
\end{table}

%% ─────────────────────────────────────────────────────────────────
\section{Prototype–Anchor Alignment and Inter-Prototype Dispersion}
\label{app:proto_dispersion}
%% ─────────────────────────────────────────────────────────────────

\begin{figure}[H]
  \centering
  \includegraphics[width=0.92\linewidth]{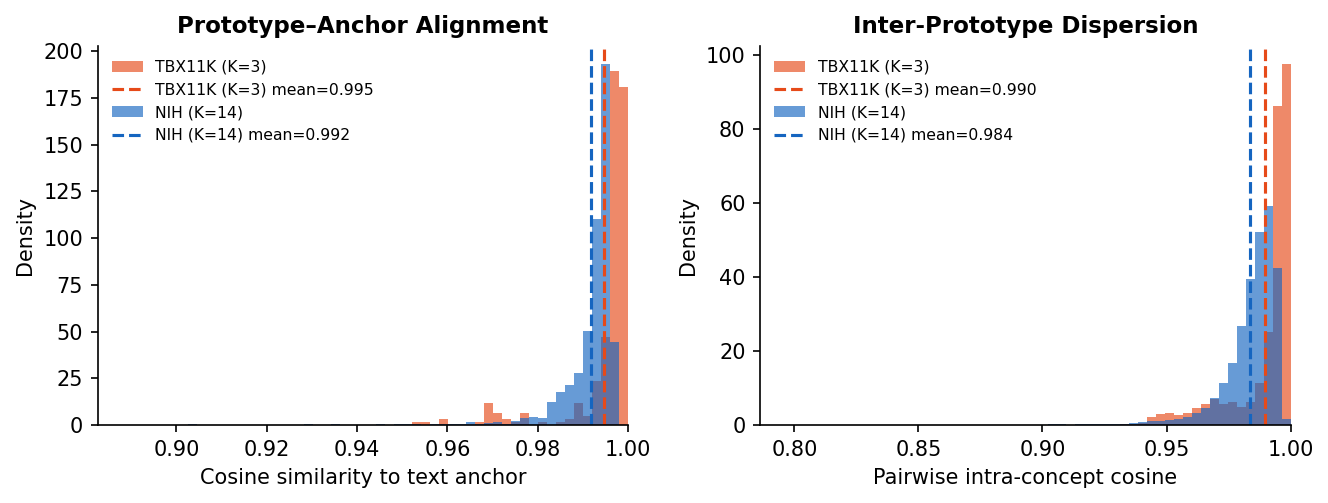}
  \caption{\textbf{Prototype geometry after Stage-3 VLM alignment.}
  \emph{Left:} distribution of cosine similarities between each prototype $\mathbf{p}_{km}$ and its text anchor $\hat{\mathbf{t}}_k$ for TBX11K ($K=3$, orange) and NIH ($K=14$, blue). Both distributions are concentrated near~1, confirming that the alignment loss ($\mathcal{L}_\text{align}$) pulls prototypes close to their anchors in both settings.
  \emph{Right:} pairwise cosine similarities between prototypes \emph{within} the same concept ($M=100$; $\binom{100}{2}=4{,}950$ pairs per concept). High intra-concept cosines indicate that the $M$ prototypes collapse toward the shared text anchor, reducing intra-concept diversity.}
  \label{fig:proto_dispersion}
\end{figure}

%% ─────────────────────────────────────────────────────────────────
\section{GAT vs.\ MLP Head Ablation}
\label{app:n6}
%% ─────────────────────────────────────────────────────────────────

\begin{table}[H]
  \caption{Ablation isolating the contribution of \emph{graph structure} vs.\ \emph{added nonlinearity} for the task head on TBX11K. A 2-layer MLP on the flattened $K{\times}M{=}300$ score vector is parameter-matched to the GAT (${\approx}47$K vs.\ ${\approx}44$K params), trained with the same Stage~4 protocol. The linear head uses the original CSR parameters (903 params). All heads share identical Stage~1--3 weights. Seed~42 single run. \textbf{Numeric note:} this table's linear-head F1 ($0.798$) differs from Table~\ref{tab:tbx11k} ($0.706$, 3-seed curriculum protocol), Table~\ref{tab:compare} ($0.728$, 5-seed), and Table~\ref{tab:graph_heads} ($0.7519$, single-seed task-head-isolation protocol) because each measures the same quantity under a different protocol (seed count, or which Stage-4 run). PG values vary across the same tables for the same reason. PG here is \textit{active\_tb} only (164 pairs).}
  \label{tab:n6_mlp}
  \centering
  \small
  \setlength{\tabcolsep}{10pt}
  \renewcommand{\arraystretch}{1.2}
  \begin{tabular}{@{}lccc@{}}
    \toprule
    Head          & Macro F1\,$\uparrow$ & PG\,$\uparrow$ & Params \\
    \midrule
    Linear (CSR)  & $0.798$              & $0.165$        & $903$     \\
    MLP (matched) & $0.865$              & $0.171$        & $47{,}235$ \\
    GAT (GRAPE)   & $\mathbf{0.896}$     & $0.165$        & $43{,}587$ \\
    \bottomrule
  \end{tabular}
  \smallskip\\
  {\footnotesize
  \textbf{Interpretation.} The MLP closes ${\sim}69\%$ of the Linear$\to$GAT F1 gap ($+6.7$\,pp out of $+9.8$\,pp total) on this single run. The remaining $+3.1$\,pp (MLP$\to$GAT) is a difference between readouts, not a graph effect, because the evaluated head's outputs do not depend on the graph at inference. PG is identical between Linear and GAT, and slightly higher for the MLP.}
\end{table}

We repeat the protocol on NIH ChestX-ray14 ($K=14$, 105 edges at $\tau=0.05$, 5 seeds \{0,1,2,3,42\}, patient-level split), retraining Stage 4 from identical frozen Stage 1--3 weights for each seed. The MLP's input grows with $K$, so on NIH it has 188K parameters vs.\ the GAT's 46K and is not parameter-matched. \emph{Hardware note.} This NIH ablation and the experiments in Appendices~\ref{app:random_anchor}, \ref{app:n13}, \ref{app:distshift}, and~\ref{app:backbone} were run on a single NVIDIA Tesla P100 16GB (batch size 32 on NIH), not the A100 80GB used for the main experiments. We do not expect the choice of GPU to affect results beyond ordinary training-run variance.

\begin{table}[H]
  \caption{GAT vs.\ MLP (188K parameters) on NIH ($K=14$), 5 seeds, patient-level split. Per-seed macro-F1 and paired $t$-test.}
  \label{tab:n6_nih}
  \centering
  \small
  \setlength{\tabcolsep}{10pt}
  \renewcommand{\arraystretch}{1.2}
  \begin{tabular}{@{}lccccc@{}}
    \toprule
    Seed & 0 & 1 & 2 & 3 & 42 \\
    \midrule
    Linear & $0.5978$ & $0.5608$ & $0.5910$ & $0.5958$ & $0.6030$ \\
    MLP (188K) & $0.5995$ & $0.5518$ & $0.5914$ & $0.5917$ & $0.5901$ \\
    GAT (GRAPE) & $0.6054$ & $0.6040$ & $0.5984$ & $0.6157$ & $0.6016$ \\
    \bottomrule
  \end{tabular}
  \smallskip\\
  \begin{tabular}{@{}lccc@{}}
    \toprule
    Head & Mean $\pm$ std & $\Delta$ vs.\ MLP & Paired $t(4)$, $p$ \\
    \midrule
    Linear & $0.5897 \pm 0.0167$ & -- & -- \\
    MLP (188K) & $0.5849 \pm 0.0189$ & -- & -- \\
    GAT (GRAPE) & $0.6050 \pm 0.0065$ & $+2.01$\,pp & $t(4)=2.33$, $p=0.080$ \\
    \bottomrule
  \end{tabular}
  \smallskip\\
  {\footnotesize
  \textbf{Interpretation.} At conventional significance ($\alpha=0.05$), the GAT does not clear the MLP by more than seed variance on NIH. No graph-specific benefit is established on either dataset, because the evaluated head's outputs do not depend on the graph at inference. The GAT did not lose to the MLP on any of the 5 seeds, and its seed-to-seed variance (std $=0.0065$) is substantially lower than the MLP's (std $=0.0189$).}
\end{table}

%% ─────────────────────────────────────────────────────────────────
\section{Self-Attention Task Head Ablation}
\label{app:n13}
%% ─────────────────────────────────────────────────────────────────

We test whether a Transformer-style self-attention head over concept nodes, which requires no predefined adjacency, outperforms the GAT. We train a parameter-matched (${\approx}48$K params) single-layer multi-head self-attention head directly on the prototype score matrix. Each concept is embedded to $d_{\text{model}}=96$, attends freely over all $K$ concepts (4 heads, no graph), and a linear readout produces class logits, replacing the GAT's fixed, training-label-derived adjacency with a fully data-driven one learned jointly with the classifier. The protocol otherwise matches Table~\ref{tab:n6_mlp}, with frozen Stage 1--3 weights (seed 42) and Stage 4 retrained on TBX11K.

\begin{table}[H]
  \caption{Self-attention vs.\ GAT task head on TBX11K, parameter-matched, identical Stage 1--3 weights, seed 42 (a separate run from Table~\ref{tab:n6_mlp}). PG is \textit{active\_tb} only (164 pairs).}
  \label{tab:n13_selfattn}
  \centering
  \small
  \setlength{\tabcolsep}{10pt}
  \renewcommand{\arraystretch}{1.2}
  \begin{tabular}{@{}lccc@{}}
    \toprule
    Head                     & Macro F1\,$\uparrow$ & PG\,$\uparrow$ & Params \\
    \midrule
    Self-attention (matched) & $\mathbf{0.906}$      & $0.085$        & $48{,}003$ \\
    GAT (GRAPE)              & $0.900$               & $\mathbf{0.091}$ & $43{,}587$ \\
    \bottomrule
  \end{tabular}
  \smallskip\\
  {\footnotesize
  \textbf{Interpretation.} Self-attention and the GAT land within seed variance of one another on both metrics, each winning narrowly on one of the two. Both are nonlinear readouts, and the self-attention head matches the GAT's performance without requiring a predefined adjacency, consistent with Appendix~\ref{app:n6}'s finding that the fixed co-occurrence graph is not responsible for the observed gain.}
\end{table}

%% ─────────────────────────────────────────────────────────────────
\section{Co-Occurrence Graph Robustness Under Distribution Shift}
\label{app:distshift}
%% ─────────────────────────────────────────────────────────────────

We test how robust the static training-label co-occurrence graph is to a shifted disease prevalence, institutional case mix, or label co-occurrence pattern. We resample the NIH training set to roughly double the prevalence of four findings (Infiltration $17.6\%\to34.1\%$, Consolidation $4.1\%\to7.6\%$, Pneumonia $1.3\%\to2.5\%$, Effusion $11.8\%\to22.6\%$), simulating a different institutional case mix, and rebuild the co-occurrence graph $\mathbf{A}$ from the resampled labels at the same threshold $\tau=0.05$. The graph changes structurally, not just numerically: 12 edges added, 2 removed. We measure macro-F1 on the original (unshifted) held-out test set, since that is what a deployed model is actually judged against, under three conditions:

\begin{table}[H]
  \caption{Macro-F1 on the original NIH test set (4,500 images) after resampling the training set to raise the prevalence of four findings and rebuilding the graph at $\tau=0.05$ (seed 42, single run).}
  \label{tab:distshift}
  \centering
  \small
  \setlength{\tabcolsep}{10pt}
  \renewcommand{\arraystretch}{1.2}
  \begin{tabular}{@{}lc@{}}
    \toprule
    Condition & Macro F1 \\
    \midrule
    Baseline (original graph and training data) & $0.6002$ \\
    Retrained under shift (new graph, Stage 4 retrained) & $0.6321$ ($+3.19$\,pp) \\
    Static-graph deployment (graph swapped only) & $0.6002$ (identical: graph unused) \\
    \bottomrule
  \end{tabular}
\end{table}

%% ─────────────────────────────────────────────────────────────────
\section{In-Domain Backbone Ablation}
\label{app:backbone}
%% ─────────────────────────────────────────────────────────────────

We test whether GRAPE's conclusions change with an in-domain foundation-model backbone instead of ImageNet-pretrained ResNet-50. Module A operates on the prototype similarity score matrix $\mathbf{S}$, not on backbone features directly, so swapping the backbone is expected to change the quality of the similarity scores Module A receives rather than what Module A does with them. The same holds for Modules B and C, which are defined over similarity maps derived from $\mathbf{S}$. We retrain the full 4-stage pipeline from scratch on TBX11K (seed 42, otherwise identical protocol) with the backbone replaced by TorchXRayVision's DenseNet-121~\cite{cohen2022torchxrayvision}, pretrained on a pool of public chest X-ray datasets (an in-domain radiology foundation model, as opposed to ImageNet-pretrained ResNet-50), producing $1024$-channel $7{\times}7$ feature maps in place of ResNet-50's $2048$-channel $7{\times}7$ maps. F1 falls by 7.5\,pp and PG rises by 5.8\,pp at the same resolution (Table~\ref{tab:backbone_ablation}). With a single seed and no re-tuning, this is not a definitive ranking of backbones.

\begin{table}[H]
  \caption{Backbone ablation on TBX11K: full four-stage retraining (GNN+Unc.+VLM, seed 42, FP16; hyperparameters not re-tuned for DenseNet-121; single run). PG over the 207 \texttt{bbox\_eval} pairs; both backbones yield $7{\times}7$ feature maps.}
  \label{tab:backbone_ablation}
  \centering
  \small
  \setlength{\tabcolsep}{10pt}
  \renewcommand{\arraystretch}{1.2}
  \begin{tabular}{@{}lccc@{}}
    \toprule
    Backbone & Macro F1\,$\uparrow$ & PG\,$\uparrow$ & Feature dim \\
    \midrule
    ResNet-50 (ImageNet) & $\mathbf{0.9001}$ & $0.0821$ & $2048$ \\
    TorchXRayVision DenseNet-121 (in-domain CXR) & $0.8255$ & $\mathbf{0.1401}$ & $1024$ \\
    \bottomrule
  \end{tabular}
\end{table}

\end{document}